\documentclass[10pt,journal,compsoc]{IEEEtran}
\PassOptionsToPackage{breaklinks=true,bookmarks=false,hyperfootnotes=false}{hyperref}

\ifCLASSOPTIONcompsoc
  \usepackage[nocompress]{cite}
\else
  \usepackage{cite}
\fi
\usepackage{times}
\usepackage{graphicx}
\usepackage{amsmath}
\usepackage{amssymb}
\usepackage{comment}
\usepackage{microtype}
\usepackage{multirow,tabularx}
\usepackage{tikz}
\usepackage{enumitem}
\usepackage[figuresleft]{rotating}

\usetikzlibrary{positioning,3d}
\usetikzlibrary{decorations.pathreplacing}

\usepackage[breaklinks=true,bookmarks=false,hyperfootnotes=false]{hyperref}

\ifCLASSINFOpdf
\else
\fi

\usepackage[caption=false,font=footnotesize,labelfont=sf,textfont=sf]{subfig}

\usepackage{stfloats}
\begin{document}
\title{Interpreting Deep Visual Representations via Network Dissection}

\author{Bolei~Zhou$^{*}$,
        David~Bau$^{*}$,
        Aude~Oliva,
        and~Antonio~Torralba% <-this % stops a space
\IEEEcompsocitemizethanks{\IEEEcompsocthanksitem B. Zhou and D.Bau contributed equally to this work. \IEEEcompsocthanksitem B. Zhou, D. Bau, A.Oliva, and A. Torralba are with CSAIL, MIT,
MA, 02139.\protect\\
E-mail: \texttt{\{bzhou, davidbau, oliva, torralba\}@csail.mit.edu}
 }% <-this % stops an unwanted space
}

\IEEEtitleabstractindextext{%
\begin{abstract}
%We propose a general framework called \textit{Network Dissection} for quantifying the interpretability of latent representations of CNNs by evaluating the alignment between individual hidden units and a set of semantic concepts. Given any CNN model, the proposed method draws on a broad dataset of visual concepts to score the semantics of hidden units at each intermediate convolutional layer. The units with semantics are given labels across a range of objects, parts, scenes, textures, materials, and colors. We use the proposed method to test the hypothesis that interpretability of units is equivalent to random linear combinations of units, then we apply our method to compare the latent representations of various networks when trained to solve different supervised and self-supervised training tasks.  We further analyze the effect of training iterations, compare networks trained with different initializations, examine the impact of network depth and width, and measure the effect of dropout and batch normalization on the interpretability of deep visual representations. We demonstrate that the proposed method can shed light on characteristics of CNN models and training methods that go beyond measurements of their discriminative power.

The success of recent deep convolutional neural networks (CNNs) depends on learning hidden representations that can summarize the important factors of variation behind the data. In this work, we describe \textit{Network Dissection}, a method that interprets networks by providing meaningful labels to their individual units. The proposed method quantifies the interpretability of CNN representations by evaluating the alignment between individual hidden units and visual semantic concepts. By identifying the best alignments, units are given interpretable labels ranging from colors, materials, textures, parts, objects and scenes. The method reveals that deep representations are more transparent and interpretable than they would be under a random equivalently powerful basis. 
We apply our approach to interpret and compare the latent representations of several network architectures trained to solve a wide range of supervised and self-supervised tasks. We then examine factors affecting the network interpretability such as the number of the training iterations, regularizations, different initialization parameters, as well as networks depth and width. Finally we show that the interpreted units can be used to provide explicit explanations of a given CNN prediction for an image. Our results highlight that interpretability is an important property of deep neural networks that provides new insights into what hierarchical structures can learn.

\end{abstract}

\begin{IEEEkeywords}
Convolutional Neural Networks, Network Interpretability, Visual Recognition, Interpretable Machine Learning.
\end{IEEEkeywords}}

\maketitle

\IEEEdisplaynontitleabstractindextext

% \fi
\IEEEpeerreviewmaketitle

\IEEEraisesectionheading{\section{Introduction}\label{sec:introduction}}


\IEEEPARstart{O}{bservations} of hidden units in deep neural networks have revealed that human-interpretable concepts can emerge as individual latent variables within those networks. For example, object detector units emerge within networks trained to recognize places \cite{zhou2014object}, part detectors emerge in object classifiers \cite{gonzalez2016semantic} and object detectors emerge in generative video networks \cite{vondrick2016generating}. This internal structure has appeared in situations where the networks are not constrained to decompose problems in any interpretable way.

The emergence of interpretable structure suggests that deep networks may be spontaneously learning disentangled representations. While a network can learn an efficient encoding that makes economical use of hidden variables to distinguish between inputs, the appearance of a disentangled representation is not well understood. A disentangled representation aligns its variables with a meaningful factorization of the underlying problem structure \cite{bengio2013representation}, or units that have a semantic interpretation (a face, wheel, green color, etc). 
%and encouraging disentangled representations is a significant area of research \cite{bengio2013representation}. 
%If the internal representation of a deep network is partly disentangled, one possible path for understanding its mechanisms is to detect disentangled structure, and read out the human interpretable factors. 
Here, we address the following key issues: 

%about the deep visual representations in this work:

{\begin{itemize}[itemsep=0mm,topsep=0mm,parsep=0mm]
\item What is a disentangled representation of neural networks, and how can its factors be detected and quantified?
\item Do interpretable hidden units reflect a special alignment of feature space?
\item What differences in network architectures, data sources, and training conditions lead to the internal representations with greater or lesser entanglement?
\end{itemize}
}

% \begin{figure}
% \begin{center}
% \input{cited-samples.tex}
% \end{center}
% \vspace{-3mm}
% \caption{Unit 13 in \cite{zhou2014object} (classifying places) detects table lamps. Unit 246 in \cite{gonzalez2016semantic} (classifying objects) detects bicycle wheels. A unit in \cite{vondrick2016generating} (self-supervised for generating videos) detects people.}
% \label{cited-figure}
% \end{figure}

We propose a general analytic framework, \textit{Network Dissection}, for interpreting deep visual representations and quantifying their interpretability. Using a broadly and densely labeled dataset named Broden, our framework identifies hidden units' semantics for any given CNN, and aligns them with interpretable concepts.

%We evaluate our method on various CNNs (AlexNet, VGG, GoogLeNet, ResNet) trained on object and scene recognition, and show that emergent interpretability is an axis-aligned property of a representation that can be destroyed by rotation without affecting discriminative power. We further examine how interpretability is affected by training datasets, training techniques like dropout \cite{srivastava2014dropout} and batch normalization \cite{ioffe2015batch}, and supervision by different primary tasks\footnote{Source code and data available at \url{http://netdissect.csail.mit.edu}}.

Building upon \cite{netdissect2017}, we provide a description of the methodology of Network Dissection in detail, and how it is used to interpret deep visual representations trained with different network architectures (AlexNet, VGG, GoogLeNet, ResNet, DenseNet) and supervisions tasks (ImageNet for object recognition, Places for scene recognition, as well as other self-taught supervision tasks). We show that interpretability is an axis-aligned property of a representation that can be destroyed by rotation without affecting discriminative power. We further examine how interpretability is affected by different training datasets, training regularizations such as dropout \cite{srivastava2014dropout} and batch normalization \cite{ioffe2015batch}, as well as fine-tuning between different data sources. Our experiments reveal that units emerge as semantic detectors in the intermediate layers of most deep visual representations, while the degree of interpretability can vary widely across changes in architecture and training sets. We conclude that representations learned by deep networks are more interpretable than previously thought, and that measurements of interpretability provide insights about the structure of deep visual representations that that are not revealed by their classification power alone\footnote{Code, data, and more dissection results are available at the project page \url{http://netdissect.csail.mit.edu/}.}.

\subsection{Related Work}

\textbf{Visualizing deep visual representations}. Though CNN models are often said to be black boxes, 
%several techniques have been developed to visualize their internal representations
their behavior can be visualized at \textit{the local individual unit level} by sampling image patches that maximize activation of hidden individual units \cite{zeiler2014visualizing,zhou2014object,girshick2016region}, or \textit{the global feature space level} by using variants of backpropagation to identify or generate salient image features \cite{mahendran2004understanding,simonyan2013deep}. Back-propagation together with a natural image prior can be used to invert a CNN layer activation \cite{mahendran2015understanding}, and an image generation network can be trained to invert the deep features by synthesizing the input images \cite{dosovitskiy2016generating}. \cite{nguyen2016synthesizing} further synthesizes the prototypical images for individual units by learning a feature code for the image generation network from \cite{dosovitskiy2016generating}.　These visualizations reveal the visual patterns that have been learned and provide a qualitative guide to unit interpretation.  In~\cite{zhou2014object}, human evaluation of visualizations is used to determine which individual units behave as object detectors in a network trained to classify scenes. However, human evaluation is not scalable to increasingly large networks such as ResNet \cite{he2016deep}. Here, we introduce a scalable method to go from qualitative visualization to quantitative interpretation of large networks. 
%In~\cite{zhou2014object}, a quantitative measure of interpretability was introduced: human evaluation of visualizations to determine which individual units behave as object detectors in a network trained to classify scenes. 

%Recently \cite{alain2016understanding} suggested an approach to testing the intermediate layers by training simple linear probes, which analyzes the information dynamics among layers and its effect on the final prediction.

%Visualizations digest the mechanisms of a network down to images which themselves must be interpreted; this motivates our work which aims to match representations of CNNs with labeled interpretations directly and automatically.

\textbf{Analyzing the properties of deep visual representations}.　 
%Various intrinsic properties of deep visual representations have been explored. 
Much work has studied the power of CNN layer activations as generic visual features for classification \cite{razavian2014cnn,agrawal2014analyzing}. While transferability of layer activations has been explored, higher layer units remain most often specialized to the target task \cite{yosinski2014transferable}.  Susceptibility to adversarial input has shown that discriminative CNN models are fooled by particular visual patterns \cite{szegedy2013intriguing,nguyen2015deep}. Analysis of correlation between different random initialized networks reveals that many units converge to the same set of representations after training \cite{li2015convergent}. The question of how representations generalize has been investigated by showing that a CNN can easily fit a random labeling of training data even under explicit regularization \cite{zhang2016understanding}. 

%Here, we focus on a less explored property of deep visual representations: their interpretability. Aude says: you say that in the next paragraph, so no need to write it twice in the same section

\textbf{Unsupervised learning of deep visual representations}. Unsupervised learning or self-supervised learning works exploit the correspondence structure that comes for free from unlabeled images to train networks from scratch \cite{doersch2015unsupervised,noroozi2016unsupervised,jayaraman2015learning,agrawal2015learning,wang2015unsupervised}. For example, a CNN is trained by predicting image context \cite{doersch2015unsupervised}, by colorizing gray images \cite{zhang2016colorful,zhang2016splitbrain}, by solving image puzzle \cite{noroozi2016unsupervised}, and by associating the images with ambient sounds \cite{owens2016ambient}. The resulting deep visual representations learned from different unsupervised learning tasks are compared by evaluating them to generic visual features on classification datasets such as Pascal VOC. Here, we provide an alternative approach to compare deep visual representations in terms of their interpretability, beyond their discriminative power.

%The deep features learned from self-supervision tasks also show promising classification performance when used for transfer learning, but their efficacy still falls short of the deep features of networks trained on millions of annotated images.

\section{Framework of Network Dissection}

The notion of a disentangled representation rests on human perception of what it means for a concept to be mixed up. Thus, we define the \textit{interpretability} of deep visual representation as the degree of alignment with human-interpretable concepts. Our quantitative measurement of interpretability proceeds in three steps:
\begin{enumerate}
\item Identify a broad set of human-labeled visual concepts.
\item Gather the response of the hidden variables to known concepts.
\item Quantify alignment of hidden variable$-$concept pairs.
\end{enumerate}
This three-step process of \emph{network dissection} is reminiscent of neuroscientists' method to characterize biological neurons~\cite{quiroga2005invariant}. Since our purpose is to measure the level to which a representation is disentangled, we focus on quantifying the correspondence between a single latent variable and a visual concept.

In a fully interpretable local coding such as a one-hot-encoding, each variable will match with one human-interpretable concept. Although we expect a network to learn partially nonlocal representations in interior layers \cite{bengio2013representation}, as past experience shows that an emergent concept will often align with a combination of a several hidden units \cite{gonzalez2016semantic,agrawal2014analyzing}, our aim is to assess how well a representation is disentangled. Therefore we measure the alignment between single units and single interpretable concepts. This does not gauge the discriminative power of the representation; rather it quantifies its disentangled interpretability. As we will show in Sec.~\ref{section-rotation}, it is possible for two representations of perfectly equivalent discriminative power to have different levels of interpretability.

To assess the interpretability of CNNs, we draw concepts from a new labeled image dataset that unifies visual concepts from a heterogeneous collection of datasets, see Sec.~\ref{section-broden}. We then measure the alignment of each CNN hidden unit with each concept by evaluating the feature activation of each individual unit as a segmentation model for each concept. To quantify the interpretability of a whole layer, we count the number of distinct concepts that are aligned with a unit, as detailed in Sec.~\ref{section-scoring}.

\subsection{Broden: Broadly and Densely Labeled Dataset}
\label{section-broden}

To ascertain alignment with both low-level concepts such as colors and higher-level concepts such as objects, we assembled the \textbf{Bro}adly and \textbf{Den}sely Labeled Dataset (\textbf{Broden}) which unifies several densely labeled image datasets: ADE \cite{zhou2016semantic}, OpenSurfaces \cite{bell14intrinsic}, Pascal-Context \cite{mottaghi_cvpr14}, Pascal-Part \cite{chen_cvpr14}, and the Describable Textures Dataset \cite{cimpoi2014describing}. These datasets contain examples of a broad range of colors, materials, textures, parts, objects and scenes. Most examples are segmented down to the pixel level except textures and scenes, which cover full images. Every pixel is also annotated automatically with one of eleven color names commonly used by humans \cite{van2009learning}. 

%Samples of the types of labels in the Broden dataset are shown in Fig.~\ref{sample_broden}. to save space, I refers to the figure in the next paragraph

\begin{figure*}
\begin{center}
\includegraphics[width=\textwidth]{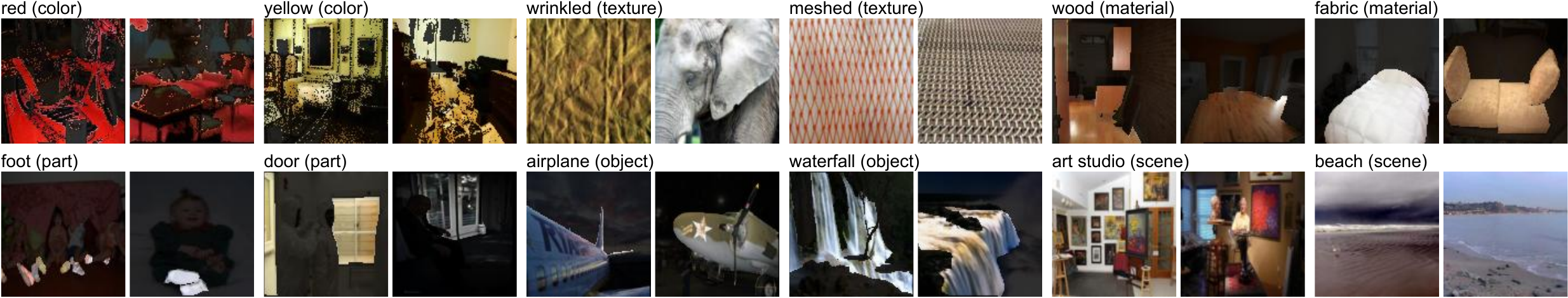}
\end{center}
\vspace{-6mm}
\caption{Samples from the Broden Dataset. The ground truth for each concept is a pixel-wise dense annotation.}\label{sample_broden}
\vspace{-4mm}
\end{figure*}

\begin{table}\caption{Statistics of each label type included in the dataset.}
\label{stat_broden}
\centering
\footnotesize
\begin{tabular}{| c | c | c | c |}
\hline
Category & Classes & Sources & Avg sample  \\
\hline
scene & 468 & ADE \cite{zhou2016semantic} & 38 \\
object & 584 & ADE \cite{zhou2016semantic}, Pascal-Context \cite{mottaghi_cvpr14} & 491 \\
part   & 234 & ADE \cite{zhou2016semantic}, Pascal-Part \cite{chen_cvpr14} & 854 \\
material & 32 & OpenSurfaces \cite{bell14intrinsic} & 1,703 \\
texture & 47 & DTD \cite{cimpoi2014describing} & 140 \\
color & 11 & Generated & 59,250 \\
\hline
\end{tabular}
\end{table}

Broden provides a ground truth set of exemplars for a set of visual concepts (see examples in Fig.~\ref{sample_broden}).  The concept labels in Broden are merged from their original datasets so that every class corresponds to an English word. Labels are merged based on shared synonyms, disregarding positional distinctions such as `left' and `top' and avoiding a blacklist of 29 overly general synonyms (such as `machine' for `car'). Multiple Broden labels can apply to the same pixel. A pixel that has the Pascal-Part label `left front cat leg' has three labels in Broden: a unified `cat' label representing cats across datasets; a similar unified `leg' label; and the color label `black'.  Only labels with at least 10 samples are included. Table~\ref{stat_broden} shows the number of classes per dataset and the average number of image samples per label class, for a total of 1197 classes.

\subsection{Scoring Unit Interpretability}
\label{section-scoring}

\begin{figure}
\begin{center}
\includegraphics[width=0.5\textwidth]{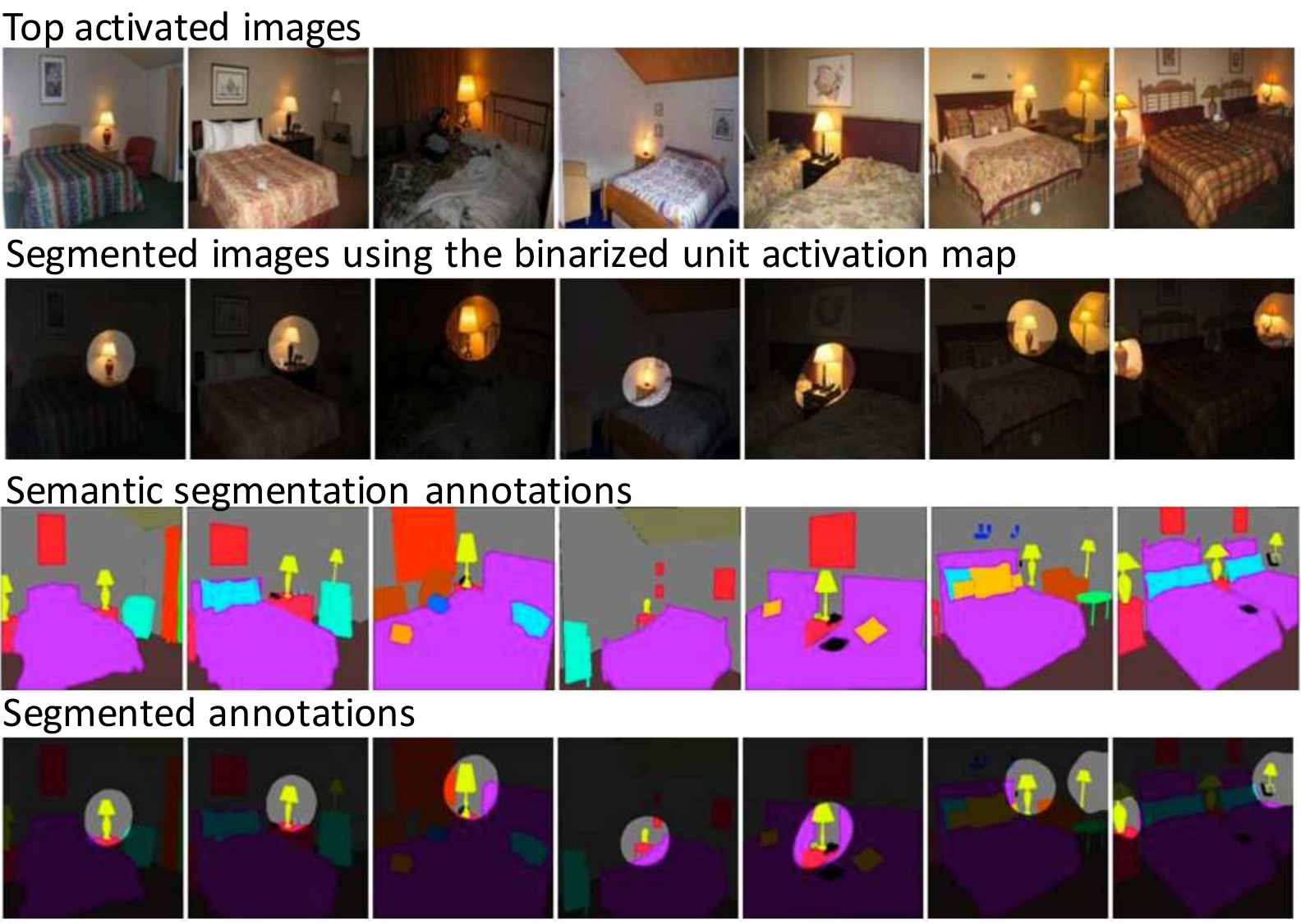}
\end{center}
\vspace{-5mm}
\caption{Scoring unit interpretability by evaluating the unit for semantic segmentation. }
\label{segmentation_mask}
\vspace{-4mm}
\end{figure}

\begin{figure*}
\begin{center}
\includegraphics[width=0.9\linewidth]{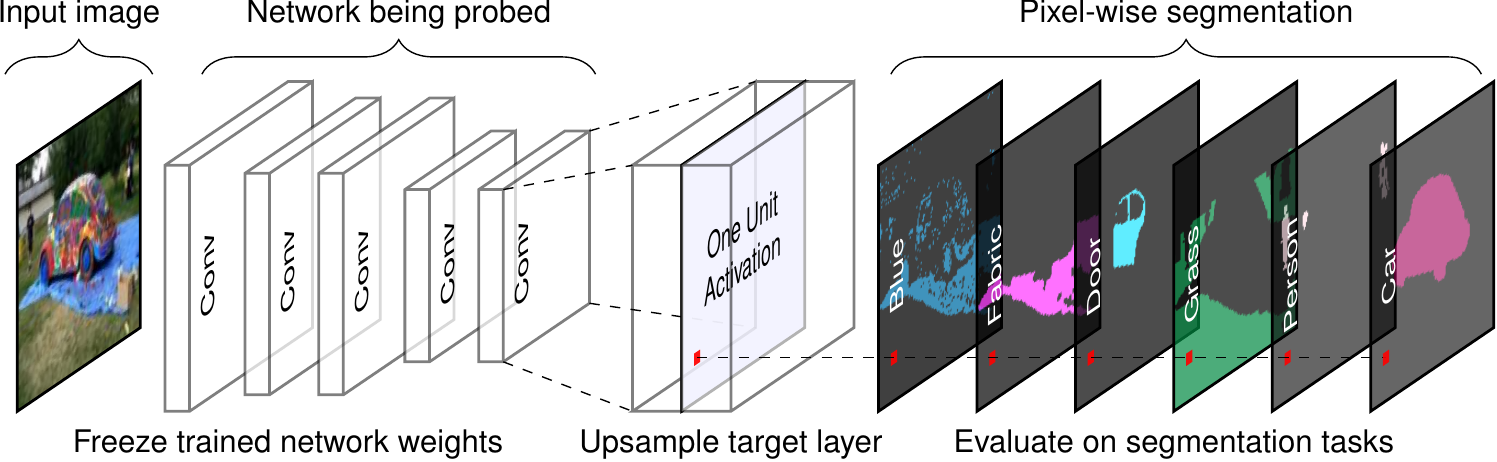}
\end{center}
\vspace{-5mm}
\caption{Illustration of network dissection for measuring semantic alignment of units in a given CNN. Here one unit of the last convolutional layer of a given CNN is probed by evaluating its performance on various segmentation tasks. Our method can probe any convolutional layer.}
\label{learning_framework}
\end{figure*}

The proposed network dissection method evaluates every individual convolutional unit in a CNN as a solution to a binary segmentation task to every visual concept in Broden (see Fig.~\ref{learning_framework}). Our method can be applied to any CNN using a forward pass without the need for training or backpropagation. For every input image $\textbf{x}$ in the Broden dataset, the activation map $A_k(\textbf{x})$ of every internal convolutional unit $k$ is collected. Then the distribution of individual unit activations $a_{k}$ is computed. For each unit $k$, the top quantile level $T_k$ is determined such that $P(a_{k} > T_k) = 0.005$ over every spatial location of the activation map in the dataset.

To compare a low-resolution unit's activation map to the input-resolution annotation mask $L_{c}$ for some concept $c$, the activation map is scaled up to the mask resolution $S_k(\textbf{x})$ from $A_k(\textbf{x})$ using bilinear interpolation, anchoring interpolants at the center of each unit's receptive field.

$S_k(\textbf{x})$ is then thresholded into a binary segmentation: $M_k(\textbf{x}) \equiv S_k(\textbf{x}) \geq T_k$, selecting all regions for which the activation exceeds the threshold $T_k$.  These segmentations are evaluated against every concept $c$ in the dataset by computing intersections $M_k(\textbf{x})\cap L_c(\textbf{x})$, for every $(k, c)$ pair.

The score of each unit $k$ as segmentation for concept $c$ is reported as a the Intersection over Union score (IoU) across all the images in the dataset,
\begin{equation}
IoU_{k,c} = \frac{\sum | M_k(\textbf{x})\cap L_c(\textbf{x})|}{\sum | M_k(\textbf{x})\cup L_c(\textbf{x})|},
\end{equation}
where $|\cdot|$ is the cardinality of a set. Because the dataset contains some types of labels which are not present on some subsets of inputs, the sums are computed only on the subset of images that have at least one labeled concept of the same category as $c$. The value of $IoU_{k,c}$ is the accuracy of unit $k$ in detecting concept $c$; we consider one unit $k$ as a \textit{detector} for concept $c$ if $IoU_{k,c}$ exceeds a threshold ($> 0.04$). Our qualitative results are insensitive to the IoU threshold: different thresholds denote different numbers of units as concept detectors across all the networks but relative orderings remain stable.  Given that one unit might be the detector for multiple concepts, here we choose the top ranked label. To quantify the interpretability of a layer, we count the number of unique concepts aligned with units, i.e. \textit{unique detectors}. 

Figure \ref{segmentation_mask} summarizes the whole process of scoring unit interpretability: By segmenting the annotation mask using the receptive field of units for the top activated images, we compute the IoU for each concept. Importantly, the IoU which evaluates the quality of the segmentation of a unit is an objective confidence score for interpretability that is \textit{comparable across networks}, enabling us to compare interpretability of different representations and so lays the basis for the experiments below. Note that network dissection results depends on the underlying vocabulary: if a unit matches a human-understandable concept that is absent from Broden, that unit will not score well for interpretability. Future versions of Broden will include a larger vocabulary of visual concepts.

\section{\textls[-25]{Experiments of Interpreting Deep Visual Representations}}
\label{sec-experiments}

\begin{table}\caption{Collection of tested CNN Models}
\label{modelzoo}
\centering
\footnotesize
\begin{tabular}{ccl}
\hline
\textbf{Training} & \textbf{Network} & \textbf{dataset or task}\\
\hline
none & AlexNet & random \\
\hline
\multirow{5}*{Supervised} & AlexNet & ImageNet, Places205, Places365, Hybrid. \\
 & GoogLeNet & ImageNet, Places205, Places365. \\
 & VGG-16 & ImageNet, Places205, Places365, Hybrid.  \\
 & ResNet-152 & ImageNet, Places365. \\
 & DenseNet-161 & ImageNet, Places365. \\
 \hline
\multirow{4}*{Self} & \multirow{4}*{AlexNet} & \texttt{context}, \texttt{puzzle}, \texttt{egomotion}, \\
&  & \texttt{tracking}, \texttt{moving}, \texttt{videoorder},\\
& &\texttt{audio}, \texttt{crosschannel},\texttt{colorization}. \\
& & \texttt{objectcentric}, \texttt{transinv}. \\
\hline
\end{tabular}
\end{table}

In this section, we conduct a series of experiments to interpret the internal representations of deep visual representations. 
In Sec.\ref{section-human}, we validate our method using human evaluation. In Sec.\ref{section-rotation} we use random unitary rotations of a learned representation to test whether interpretability of CNNs is an axis-independent property; we find that it is not, and we conclude that interpretability is not an inevitable result of the discriminative power of a representation.  In Sec.\ref{section-comparing-architecture} we analyze all the convolutional layers of AlexNet as trained on ImageNet \cite{krizhevsky2012imagenet} and Places \cite{zhou2014learning}. We confirm that our method reveals detectors for higher-level semantic concepts at higher layers and lower-level concepts at lower layers; and that more detectors for higher-level concepts emerge under scene training. Then, we show that different network architectures such as AlexNet, VGG, and ResNet yield different interpretability, and differently supervised training tasks and self-supervised training tasks also yield a variety of levels of interpretability in Sec.\ref{section-selfsupervision}. Additionally in Sec.\ref{section-captioning} we show the interpretability of model trained from captioning images. Another set of experiments shows the impact of different training conditions in Sec.\ref{trainingcondition} and what happens during the transfer learning in Sec.\ref{section-transfer-learning}. We further examine the relationship between discriminative power and interpretability in Sec.\ref{section-discrimination}, and investigate a possible way to improve the interpretability of CNNs by increasing their width in Sec.\ref{section-layerwidth}. Finally in Sec.\ref{section-explaination}, we utilize the interpretable units as explanatory factors to the prediction given by a CNN.

For testing we used CNN models with different architectures and primary tasks (Table \ref{modelzoo}), including AlexNet \cite{krizhevsky2012imagenet}, GoogLeNet \cite{szegedy2015going}, VGG \cite{simonyan2014very}, ResNet \cite{he2016deep}, and DenseNet \cite{huang2016densely}. For supervised training, the models are trained from scratch (i.e., not pretrained) on ImageNet \cite{russakovsky2015imagenet}, Places205 \cite{zhou2014learning}, and Places365 \cite{zhou2016places}. ImageNet is an object-centric dataset, which contains 1.2 million images from 1000 object classes. Places205 (2.4 million images from 205 scene classes) and Places365 (1.6 million images from 365 scene classes) are two subsets the scene-centric dataset Places.``Hybrid'' network refers to a combination of ImageNet and Places365. The self-supervised networks are introduced in Sec.\ref{section-selfsupervision}.

\subsection{Human Evaluation of Interpretations}
\label{section-human}

Using network dissection, we analyzed the interpretability of units within all the convolutional layers of Places-AlexNet and ImageNet-AlexNet, then compared with human interpretation. Places-AlexNet is trained for scene classification on Places205 \cite{zhou2014learning}, while ImageNet-AlexNet is the identical architecture trained for object classification on ImageNet \cite{krizhevsky2012imagenet}.

\begin{figure*}
\begin{center}
\includegraphics{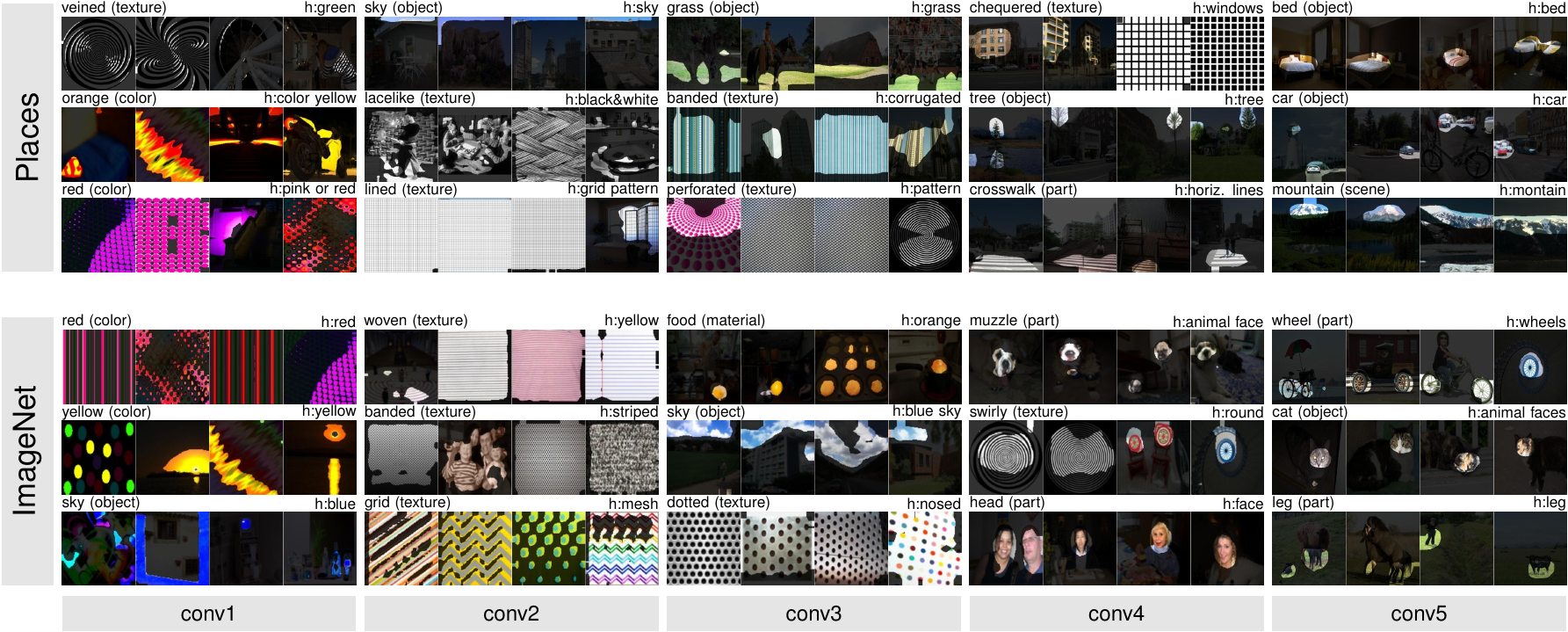}
\end{center}
\vspace{-5mm}
\caption{Comparison of the interpretability of the convolutional layers of AlexNet, trained on classification tasks for Places (top) and ImageNet (bottom).Four units in each layer are shown with their semantics. The segmentation generated by each unit is shown on the three Broden images with highest activation. Top-scoring labels are shown above to the left, and human-annotated labels are shown above to the right. There is some disagreement: for example, raters mark the first \texttt{conv4} unit on Places as a `windows' detector, while the algorithm matches the `chequered' texture.}\label{places_imagenet}
\end{figure*}

Our evaluation was done by raters on Amazon Mechanical Turk (AMT). As a baseline, we used the descriptions from \cite{zhou2014object}, where three independent raters wrote short phrases and gave a confidence score, to describe the meaning of a unit, based on seeing the top image patches for that unit. As a canonical description. we chose the most common description of a unit (when raters agreed), and the highest-confidence description (when raters did not agree). To identify non-interpretable units, raters were shown the canonical descriptions of visualizations and asked whether the description was valid.  Units with validated descriptions are taken as interpretable units. To compare these baseline descriptions to network-dissection-derived labels, raters were shown a visualization of top images patches for an interpretable unit, along with a word or short phrase, and asked to vote (yes/no) whether the phrase was descriptive of most of the patches.  The baseline human-written descriptions were randomized with the labels from net dissection, and the origin of the labels was not revealed to the raters. Table~\ref{comparison_quantitative} summarizes the results.  The number of interpretable units is shown for each layer and type of description.  As expected, color and texture concepts dominate in the lower layers \texttt{conv1} and \texttt{conv2} while part, object and scene detectors are more frequent at \texttt{conv4} and \texttt{conv5}.  Average positive votes for descriptions of interpretable units are shown, both for human-written labels and network-dissection-derived labels. Human labels are most highly consistent for units of \texttt{conv5}, suggesting that humans have no trouble identifying high-level visual concept detectors, while lower-level detectors, particularly textures, are more difficult to label.  Similarly, labels given by network dissection are best at \texttt{conv5} and for high-level concepts, and are found to be less descriptive for lower layers and textures. In Fig.~\ref{places_imagenet}, a sample of units is shown together with both automatically inferred interpretations and manually assigned interpretations taken from \cite{zhou2014object}. The predicted labels match the human annotation well, though sometimes they capture a different description of a concept, like the `crosswalk' predicted by the algorithm compared to `horizontal lines' given by human for the third unit in \texttt{conv4} of Places-AlexNet in Fig.~\ref{places_imagenet}.

\begin{figure}
\centering
{%
\setlength{\fboxsep}{0.03\columnwidth}%
\setlength{\fboxrule}{1pt}%
\fbox{\includegraphics[width=0.7\columnwidth]{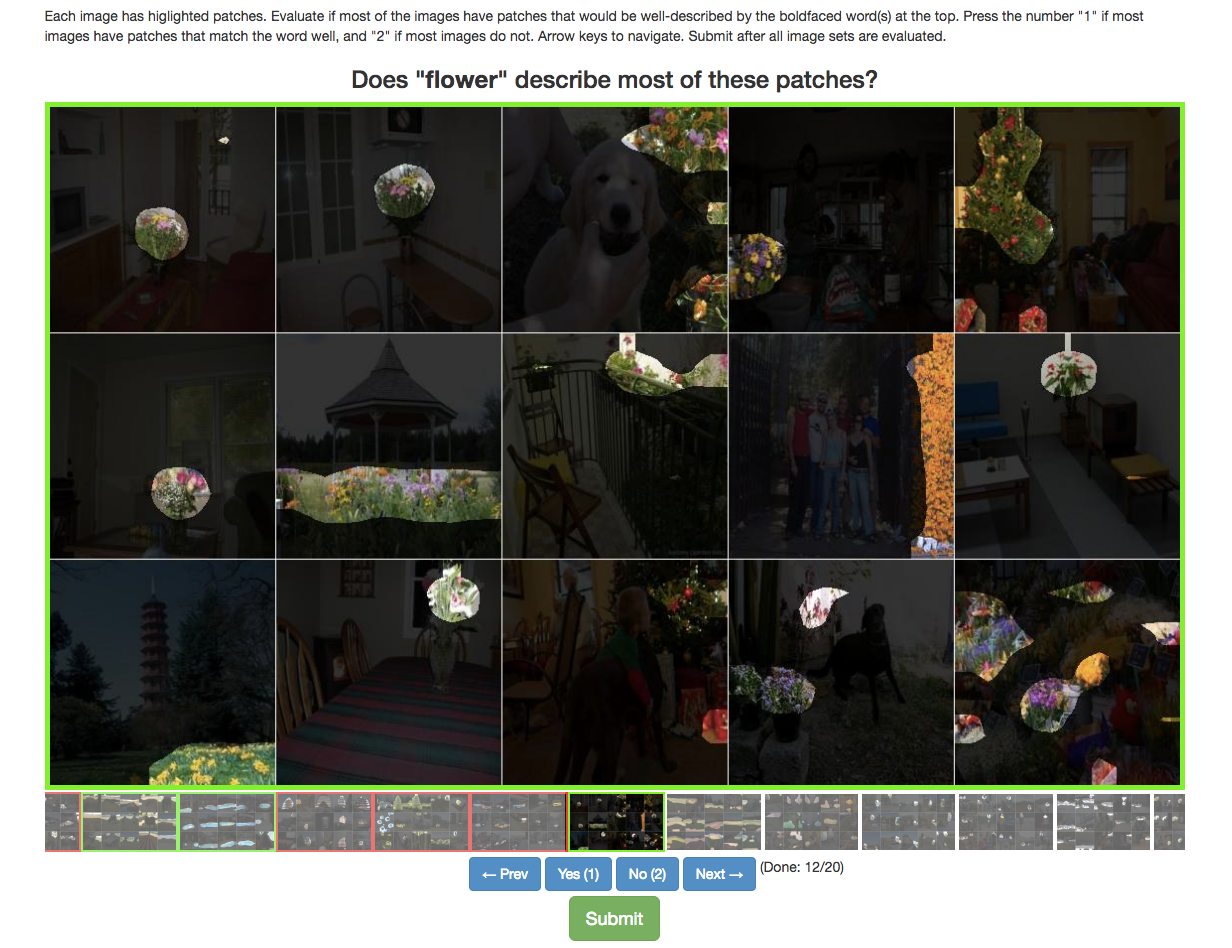}}%
}%
\caption{The annotation interface used by human raters on Amazon Mechanical Turk. Raters are shown descriptive text in quotes together with fifteen images, each with highlighted patches, and must evaluate whether the quoted text is a good description for the highlighted patches.}\label{amt_interface}
\end{figure}

\begin{table}\caption{Human evaluation of our Network Dissection approach.}
\label{comparison_quantitative}
\centering
\footnotesize
\begin{tabular}{@{\hspace{1mm}}lll@{\hspace{2.2mm}}l@{\hspace{2.2mm}}l@{\hspace{2.2mm}}l@{\hspace{1mm}}}
\hline
 & conv1 & conv2 & conv3 & conv4 & conv5 \\
\hline
Interpretable units & 57/96 & 126/256 & 247/384 & 258/384 & 194/256 \\
color units & 36 & 45 & 44 & 19 & 12 \\
texture units & 19 & 53 & 64 & 72 & 23 \\
material units & 0 & 2 & 2 & 9 & 8 \\
part units & 0 & 0 & 13 & 17 & 16 \\
object units & 2 & 22 & 109 & 127 & 114 \\
scene units & 0 & 4 & 15 & 14 & 21 \\
\hline
Human consistency & 82\% & 76\% & 83\% & 82\% & 91\% \\
on color units & 92\% & 80\% & 82\% & 84\% & 100\% \\
on texture units & 68\% & 81\% & 83\% & 81\% & 96\% \\
on material units & n/a & 50\% & 100\% & 78\% & 100\% \\
on part units & n/a & n/a & 92\% & 94\% & 88\% \\
on object units & 50\% & 68\% & 84\% & 83\% & 90\% \\
on scene units & n/a & 25\% & 67\% & 71\% & 81\% \\
\hline
Network Dissection & 37\% & 56\% & 54\% & 59\% & 71\% \\
on color units & 44\% & 53\% & 55\% & 42\% & 67\% \\
on texture units & 26\% & 58\% & 42\% & 54\% & 39\% \\
on material units & n/a & 50\% & 50\% & 89\% & 75\% \\
on part units & n/a & n/a & 85\% & 71\% & 75\% \\
on object units & 0\% & 59\% & 57\% & 65\% & 75\% \\
on scene units & n/a & 50\% & 53\% & 29\% & 86\% \\
\hline
\end{tabular}
\end{table}

%Using network dissection, we analyze and compare the interpretability of units within all the convolutional layers of Places-CNN. Places-AlexNet is trained for scene classification on Places205 \cite{zhou2014learning}, while ImageNet-AlexNet is the identical architecture trained for object classification on ImageNet \cite{krizhevsky2012imagenet}.

\subsection{Measurement of Axis-aligned Interpretability}
\label{section-rotation}

Two hypotheses can explain the emergence of interpretability in individual hidden layer units:
\begin{enumerate}[itemindent=4em]
\item[Hypothesis 1.] Interpretability is a property of the representation as a whole, and individual interpretable units emerge because interpretability is a generic property of typical directions of representational space. Under this hypothesis, projecting to \textit{any} direction would typically reveal an interpretable concept, and interpretations of single units in the natural basis would not be more meaningful than interpretations that can be found in any other direction.
\item[Hypothesis 2.] Interpretable alignments are unusual, and interpretable units emerge because learning converges to a special basis that aligns explanatory factors with individual units. In this model, the natural basis represents a meaningful decomposition learned by the network.
\end{enumerate}
Hypothesis 1 is the default assumption: in the past it has been found~\cite{szegedy2013intriguing} that with respect to interpretability ``there is no distinction between individual high level units and random linear combinations of high level units.'' Network dissection allows us to re-evaluate this hypothesis. Thus, we conduct an experiment to determine whether it is meaningful to assign an interpretable concept to an individual unit. We apply random changes in basis to a representation learned by AlexNet. Under hypothesis 1, the overall level of interpretability should not be affected by a change in basis, even as rotations cause the specific set of represented concepts to change. Under hypothesis 2, the overall level of interpretability is expected to drop under a change in basis.

\begin{figure}
\begin{center}
\vspace{-5mm}
\includegraphics[width=0.9\linewidth]{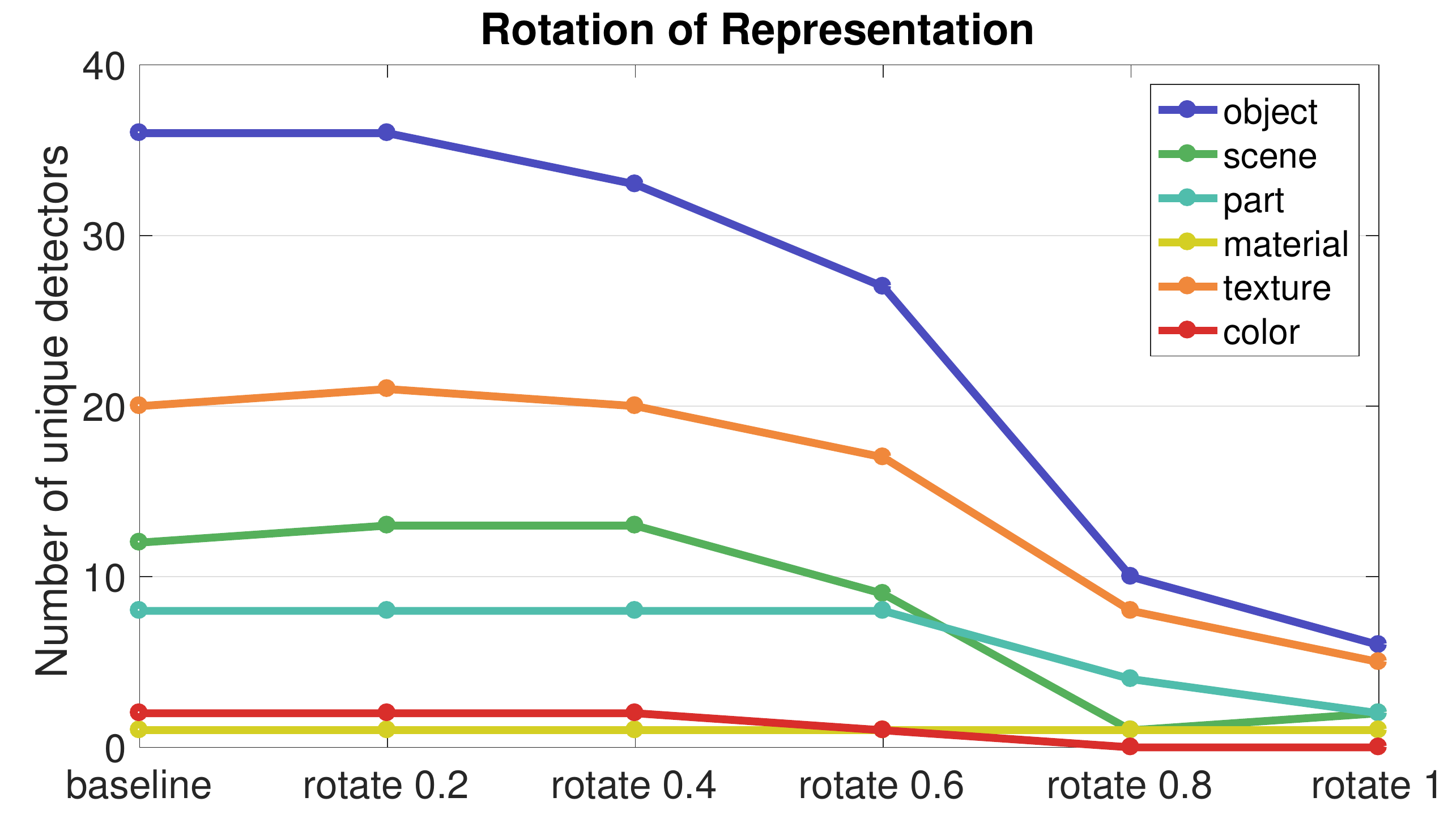}
\end{center}
\vspace{-6mm}
\caption{Interpretability over changes in basis of the representation of AlexNet \texttt{conv5} trained on Places. The vertical axis shows the number of unique interpretable concepts that match a unit in the representation.  The horizontal axis shows $\alpha$, which quantifies the degree of rotation.}
\label{rotation-plot}
\end{figure}

We begin with the representation of the 256 convolutional units of AlexNet \texttt{conv5} trained on Places205 and examine the effect of a change in basis. To avoid any issues of conditioning or degeneracy, we change basis using a random orthogonal transformation $Q$. The rotation $Q$ is drawn uniformly from $SO(256)$ by applying Gram-Schmidt on a normally-distributed $QR = A \in \mathbf{R}^{256^2}$ with positive-diagonal right-triangular $R$, as described by \cite{diaconis2005random}. Interpretability is summarized as the number of unique visual concepts aligned with units, as defined in Sec.~\ref{section-scoring}.

Denoting AlexNet \texttt{conv5} as $f(x)$, we found that the number of unique detectors in $Qf(x)$ is 80\% fewer than the number of unique detectors in $f(x)$. Our finding is inconsistent with hypothesis 1 and consistent with hypothesis 2.

\begin{figure}
\begin{center}
\includegraphics[width=1\linewidth]{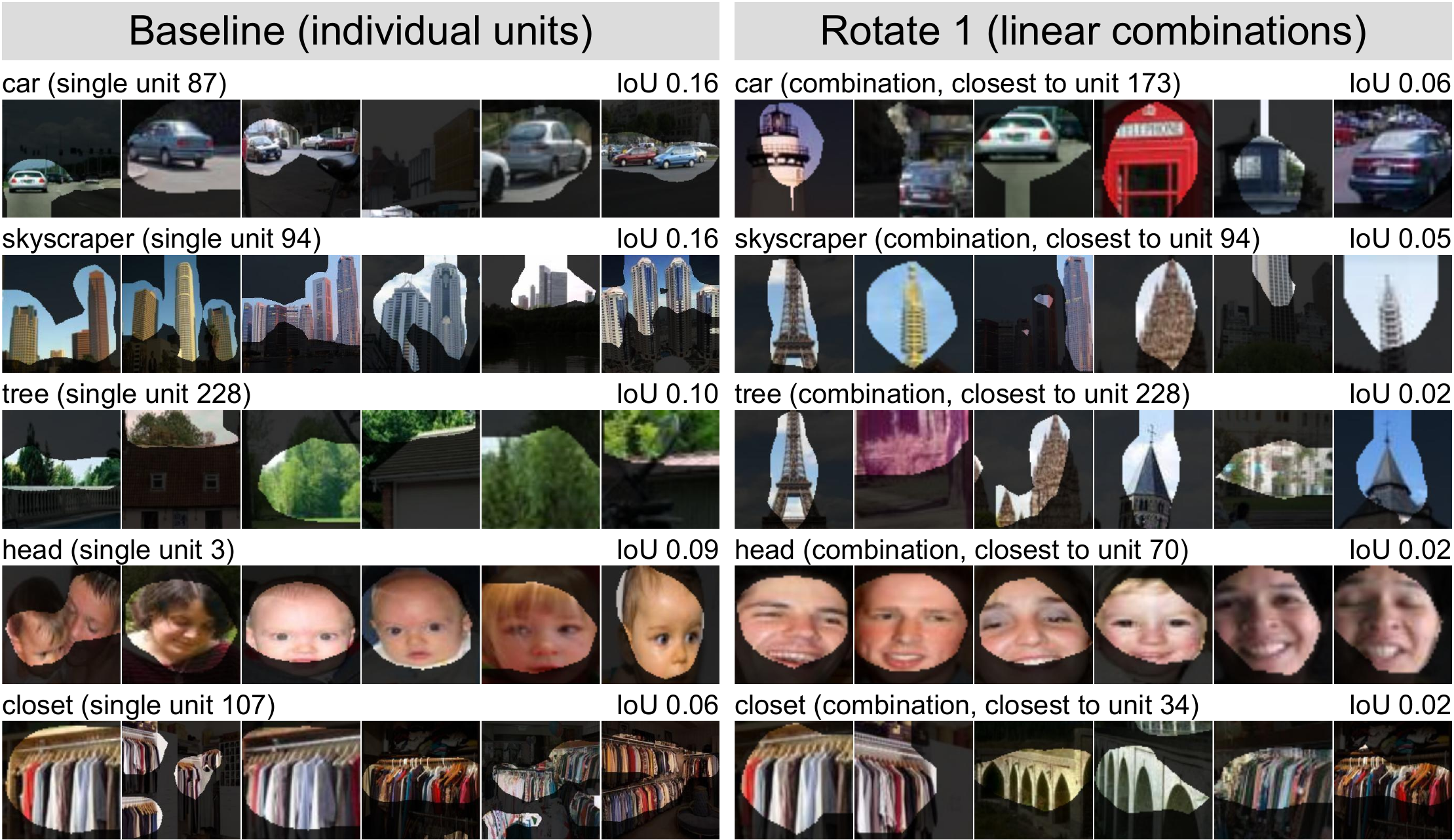}\end{center}
\vspace{-5mm}
\caption{Visualizations of the best single-unit concept detectors of five concepts taken from individual units of AlexNet \texttt{conv5} trained on Places (left), compared with the best linear-combination detectors of the same concepts taken from the same representation under a random rotation (right).  For most concepts, both the IoU and the visualization of the top activating image patches confirm that individual units match concepts better than linear combinations. In other cases, (e.g. head detectors) visualization of a linear combination appears highly consistent, but the IoU reveals lower consistency when evaluated over the whole dataset.}
\label{rotation-qualitative}
\vspace{-4mm}
\end{figure}

We also tested smaller perturbations of basis using $Q^{\alpha}$ for $0 \leq \alpha \leq 1$,
where the fractional powers $Q^\alpha \in SO(256)$ are chosen to form a minimal geodesic gradually rotating from $I$ to $Q$; these intermediate rotations are computed using a Schur decomposition. Fig.~\ref{rotation-plot} shows that interpretability of $Q^\alpha f(x)$ decreases as larger rotations are applied. Fig.~\ref{rotation-qualitative} shows some examples of the linearly combined units.

Each rotated representation has the same discriminative power as the original layer. Writing the original network as $g(f(x))$, note that $g'(r) \equiv g(Q^T r)$ defines a neural network that processes the rotated representation $r = Q f(x)$ exactly as the original $g$ operates on $f(x)$.  Furthermore, we verify that a network can learn to solve a task given a rotated representation.  Starting with AlexNet trained to solve places365, we freeze the bottom layers up to $\texttt{pool5}$ and retrain the top layers of the network under two conditions: one in which the representation at \texttt{pool5} is randomly rotated  ($\alpha=1$) before passing to \texttt{fc6}, and the other where the representation up to $\texttt{pool5}$ is left unchanged.  Then we reinitialize and retrain the \texttt{fc6}-\texttt{fc8} layers of an AlexNet on places365.  Under both the unrotated and rotated conditions, reinitializing and retraining the top layers improves performance, and the improvement is similar regardless of whether the \texttt{pool5} representation is rotated.  Initial accuracy is 50.3\%.  After retraining the unrotated representation, accuracy improves to 51.9\%; after retraining the rotated representation, accuracy is 51.7\%. Thus the network learns to solve the task even when the representation is randomly rotated.  Since a network can be transformed into an equivalent network with the same discriminative ability but with lower interpretability, we conclude that interpretability must be measured separately from discrimination ability.

\begin{figure}
\begin{center}
\includegraphics[width=0.9\linewidth]{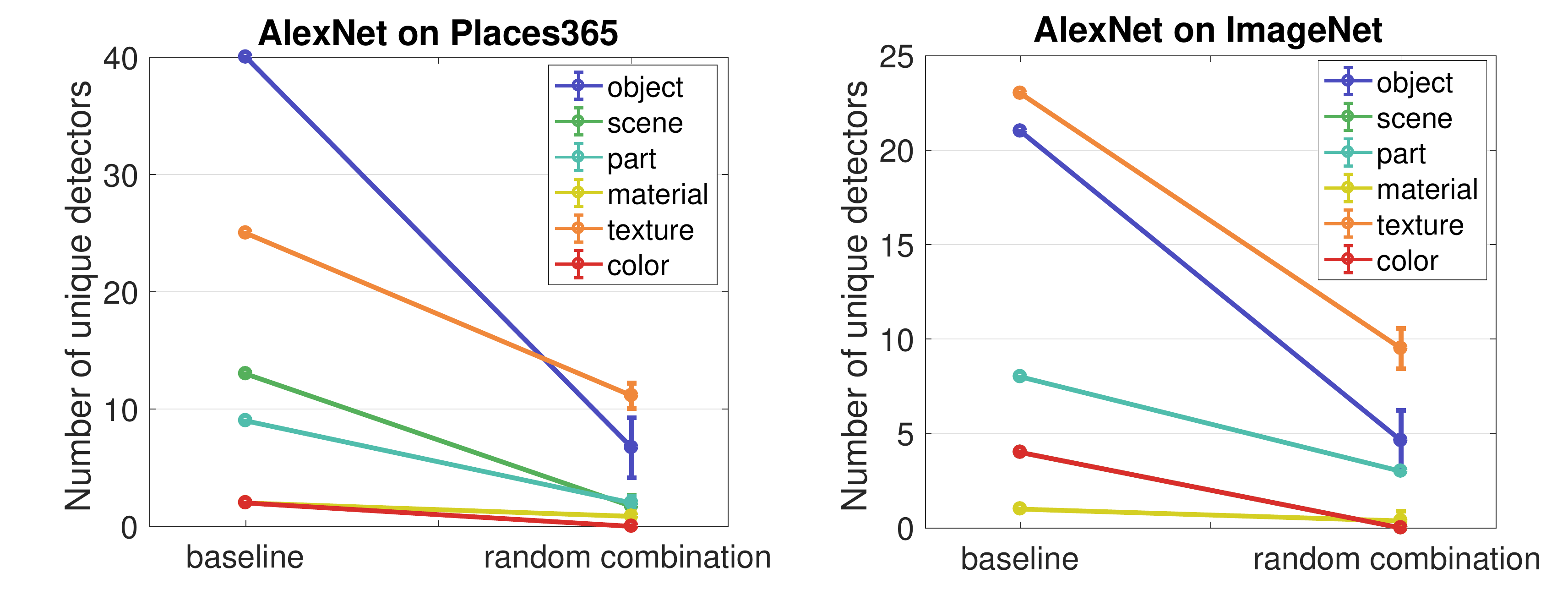}
\end{center}
\vspace{-6mm}
\caption{Complete rotation ($\alpha=1$) repeated on AlexNet trained on Places365 and ImageNet respectively. Rotation reduces the interpretability significantly for both of the networks.}
\label{rotation-repeat}
\end{figure}

We repeated the measurement of interpretability upon complete rotation ($\alpha=1$) on Places365 and ImageNet 10 times; see results in Fig. \ref{rotation-repeat}. There is a drop of interpretability for both.  Alexnet on Places365 drops more, which can be explained due to that network starting with a higher number of interpretable units.

\begin{figure}
\begin{center}
\includegraphics[width=1\linewidth]{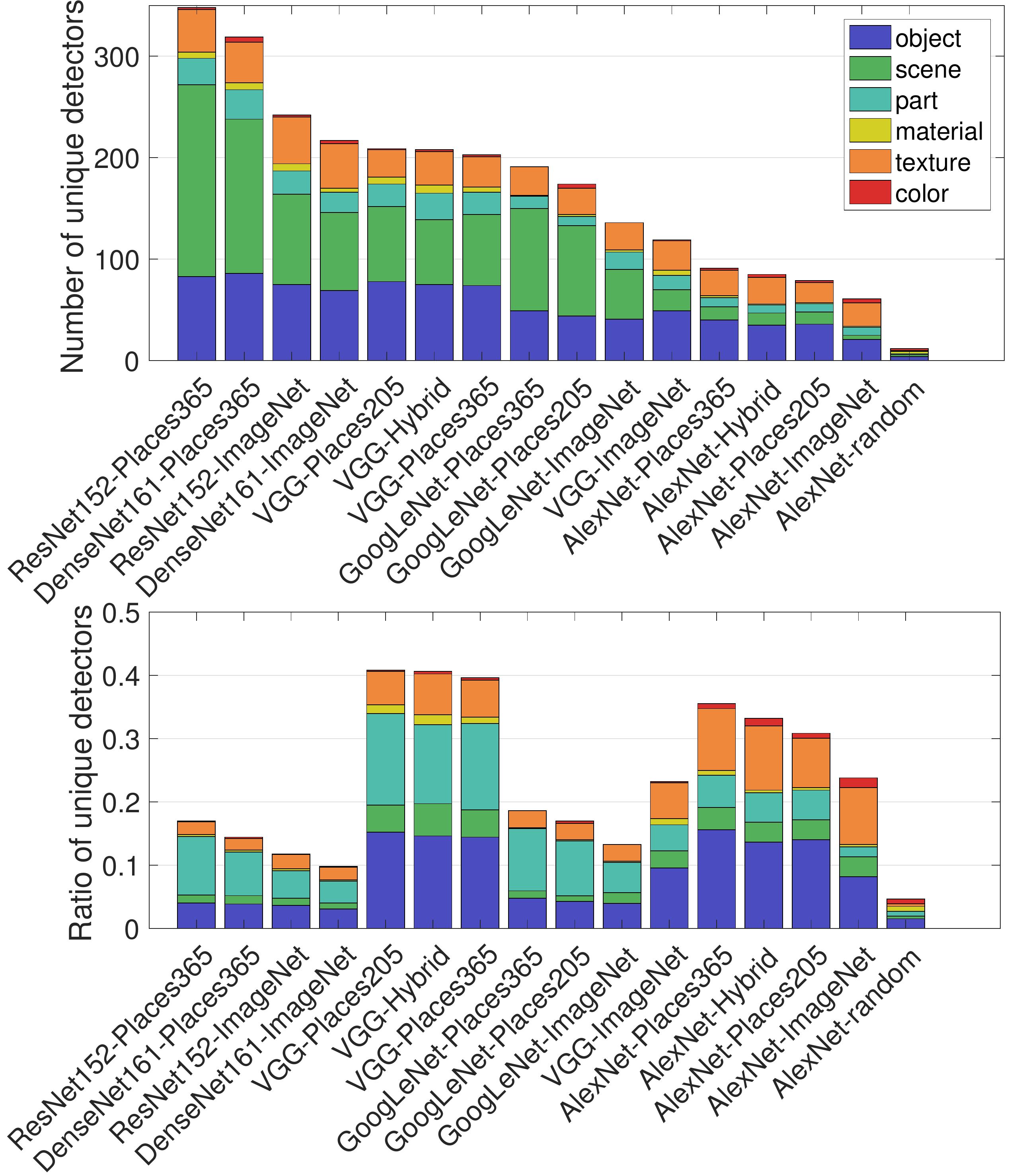}
\end{center}
\vspace{-5mm}
\caption{Interpretability across different architectures trained on ImageNet and Places. Plot above shows the number of unique detectors, plot below shows the ratio of unique detectors (number of unique detectors divided by the total number of units).}\label{architecture}
\end{figure}

\subsection{Network Architectures with Supervised Learning}
\label{section-comparing-architecture}

% \begin{figure}
% \begin{center}
% \includegraphics[width=0.5\textwidth]{plot/all_histograms.pdf}
% \end{center}
% \caption{Histogram of detectors for the AlexNet-Places365.}
% \label{figure-histogram-baseline}
% \end{figure}

How do different network architectures affect disentangled interpretability of the learned representations? For simplicity, the following experiments focus on the last convolutional layer of each CNN, where semantic detectors emerge most.
%We apply network dissection to evaluate several network architectures trained on ImageNet and Places. 

Results showing the number of unique detectors that emerge from various network architectures trained on ImageNet and Places, , and the ratio of unique detectors (the number of unique detectors normalized by the total number of units at that layer) are plotted in Fig.~\ref{architecture}. Interpretability in terms of the number of unique detectors, can be compared as follows: ResNet $>$ DenseNet $>$ VGG $>$ GoogLeNet $>$ AlexNet. Deeper architectures seem to have greater interpretability, though individual layer structure is different across architectures. Comparing training datasets, we find Places $>$ ImageNet. As discussed in \cite{zhou2014object}, scenes are composed of multiple objects, with more object detectors emerging in CNNs trained to recognize places. In terms of ratio of unique detectors, VGG architecture is highest. We consider the number of unique detectors as the metric of interpretability for a network as it better measures the diversity and coverage of emergent interpretable concepts.

% \begin{figure}
% \begin{center}
% \includegraphics[width=1\linewidth]{plot/detail-study-crop.pdf}
% \end{center}
% \vspace{-6mm}
% \caption{Histogram of the object detectors from the ResNet and DenseNet trained on ImageNet and Places respectively.}
% \label{hist_objectdetector}
% \end{figure}

\begin{figure}
\begin{center}
\includegraphics[width=1\linewidth]{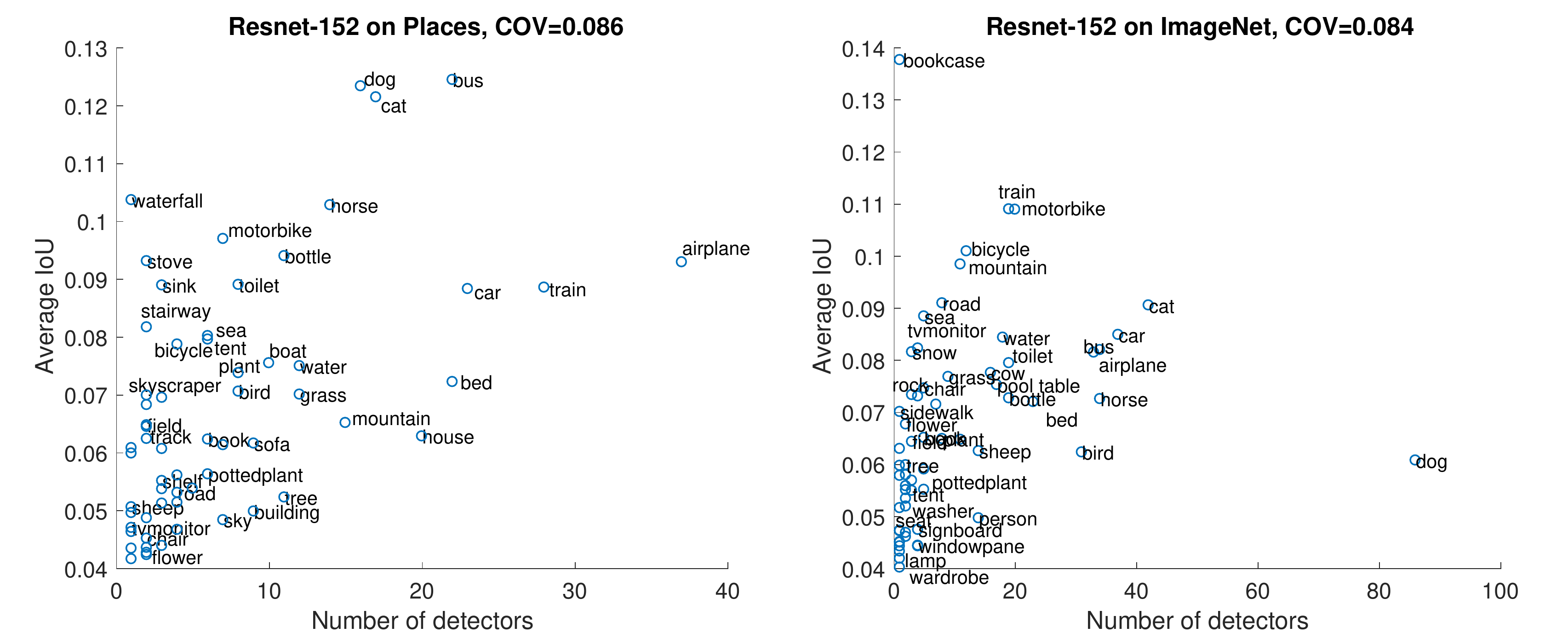}
\end{center}
\vspace{-6mm}
\caption{Average IoU versus the number of detectors for the object class in Resnet152 trained on Places and ImageNet respectively. For a set of units detecting the same object class, we average their IoU. }
\label{averageIoU}
\end{figure}

\begin{figure*}
\begin{center}
\includegraphics[width=1\linewidth]{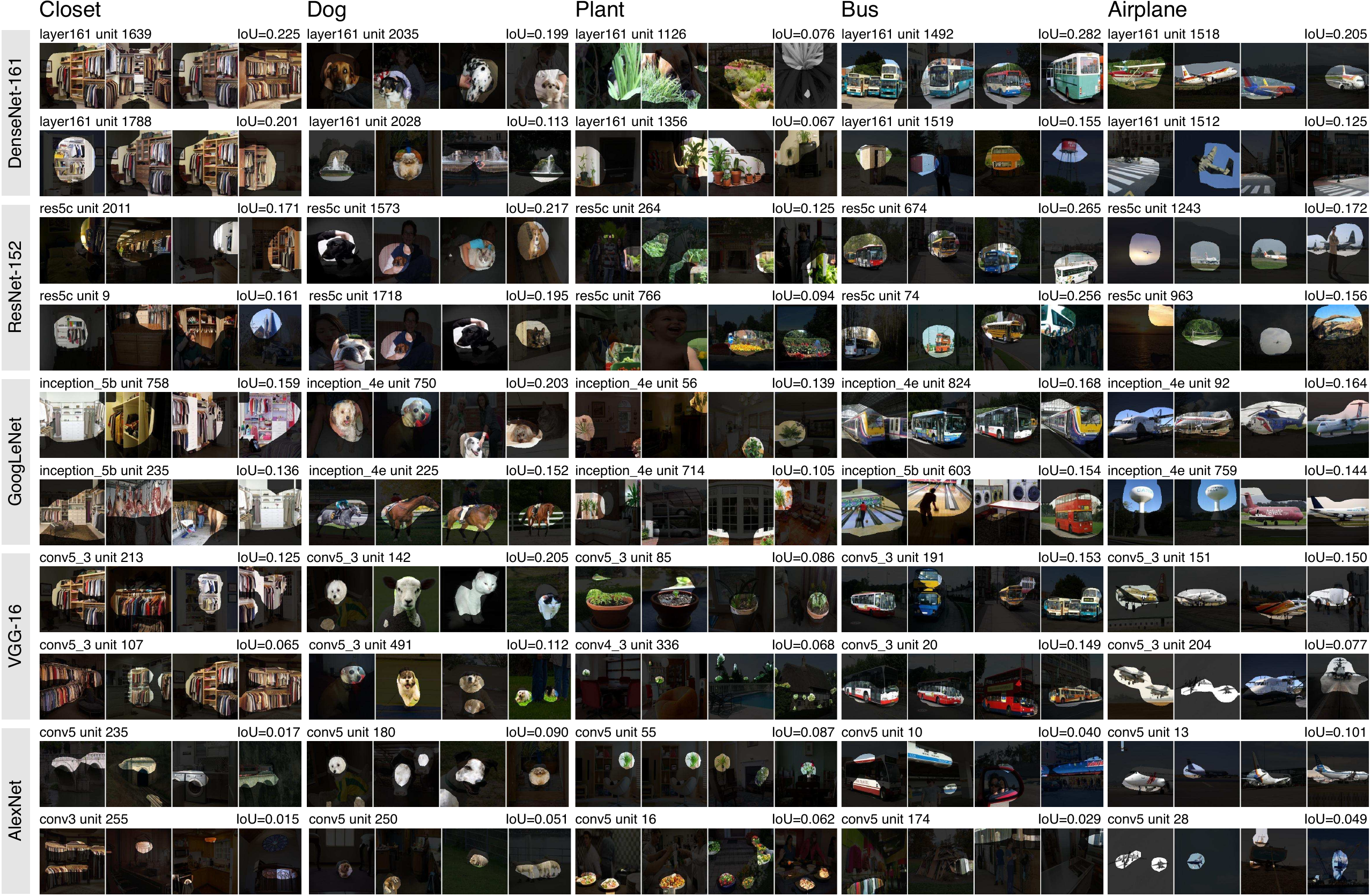}
\end{center}
\vspace{-5mm}
\caption{Comparison of several visual concept detectors identified by network dissection in DenseNet, ResNet, GoogLeNet, VGG, and AlexNet. Each network is trained on Places365. The two highest-IoU matches among convolutional units of each network is shown. The segmentation generated by each unit is shown on the four maximally activating Broden images. Some units activate on concept generalizations, e.g., GoogLeNet 4e's unit 225 on horses and dogs, and 759 on white ellipsoids and jets.}
\label{figure-concepts}
\vspace{-3mm}
\end{figure*}

%\begin{sidewaysfigure*}
%\begin{center}
%\includegraphics[width=0.95\linewidth]{plot/full-study-split-crop.pdf}
%\end{center}
%\caption{Comparison of unique detectors of all types on a variety of architectures. More results are at the project page.\textcolor{red}{make this figure only occupy 1/4 page. so maybe just include various alexnet architectures.}}
%\label{sidewayfigure_all}
%\end{sidewaysfigure*}

\begin{figure}
\begin{center}
\includegraphics[width=0.95\linewidth]{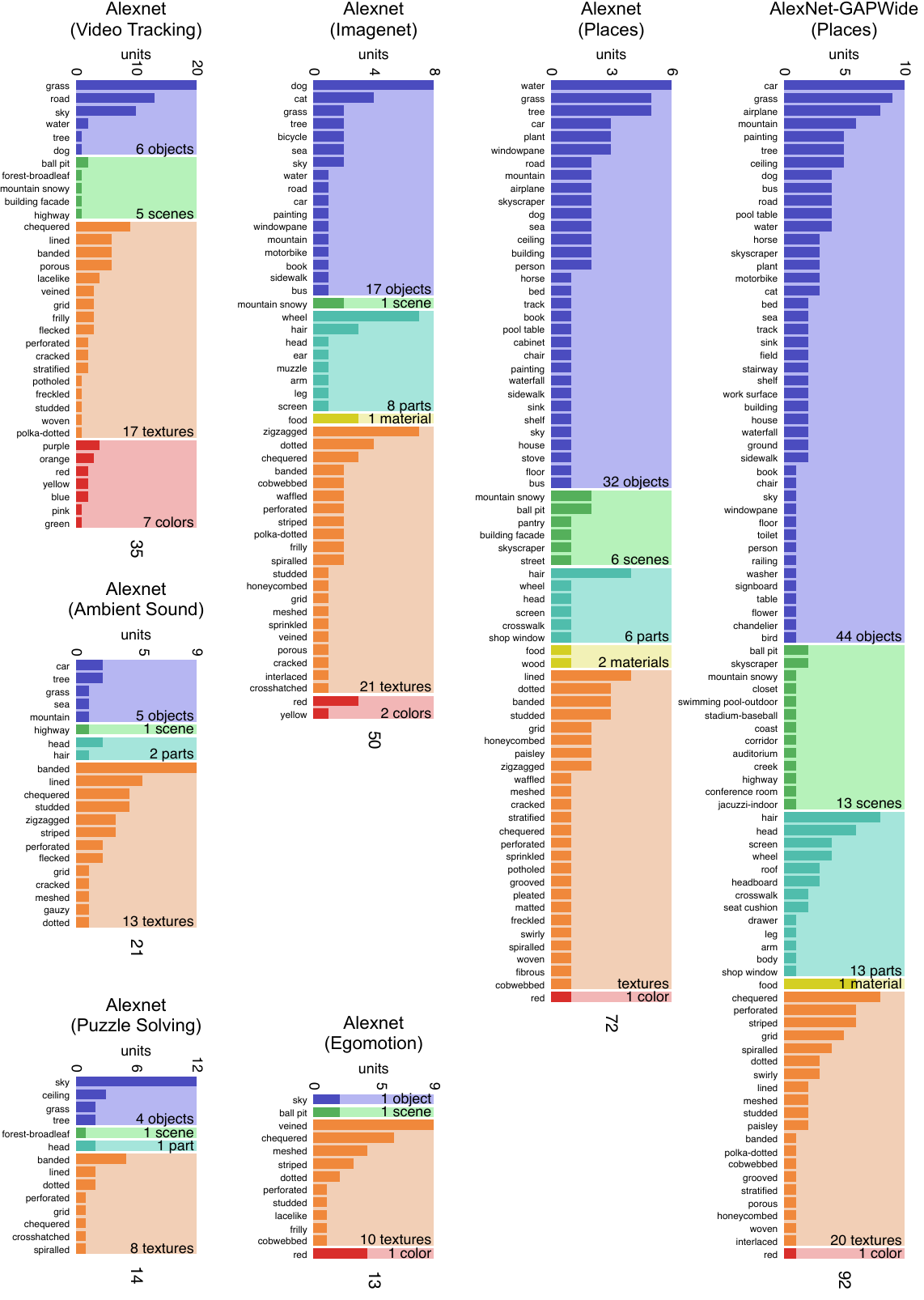}
\end{center}
\caption{Comparison of unique detectors of all types on a variety of training tasks. More results, including comparisons across architectures, are at the project page.}
\label{bargraph_comparison}
\end{figure}

%Fig.~\ref{hist_objectdetector} shows the histogram of object detectors identified inside ResNet and DenseNet trained on ImageNet and Places respectively. DenseNet161-Places365 has the largest number of unique object detectors among all the networks. The emergent detectors differ across both training data source and architecture. The most frequent object detectors in the two networks trained on ImageNet are dog detectors, given the over 100 dog categories in the ImageNet training set. 

Fig.~\ref{averageIoU} shows the plot of average IoU versus the number of detectors for the object detectors in Resnet152 trained on Places and ImageNet. Note the weak positive correlation between the two (r=0.08), i.e, the higher average IoU the more detectors for that class.

Fig.~\ref{figure-concepts} shows some object detectors grouped by object categories. For the same object category, the visual appearance of the unit as detector varies within the same network and across different networks. DenseNet and ResNet have such good detectors for bus and airplane with IoU $> 0.25$. Fig.~\ref{bargraph_comparison} compares interpretable units on a variety of training tasks.

\begin{figure*}
\includegraphics[width=1\textwidth]{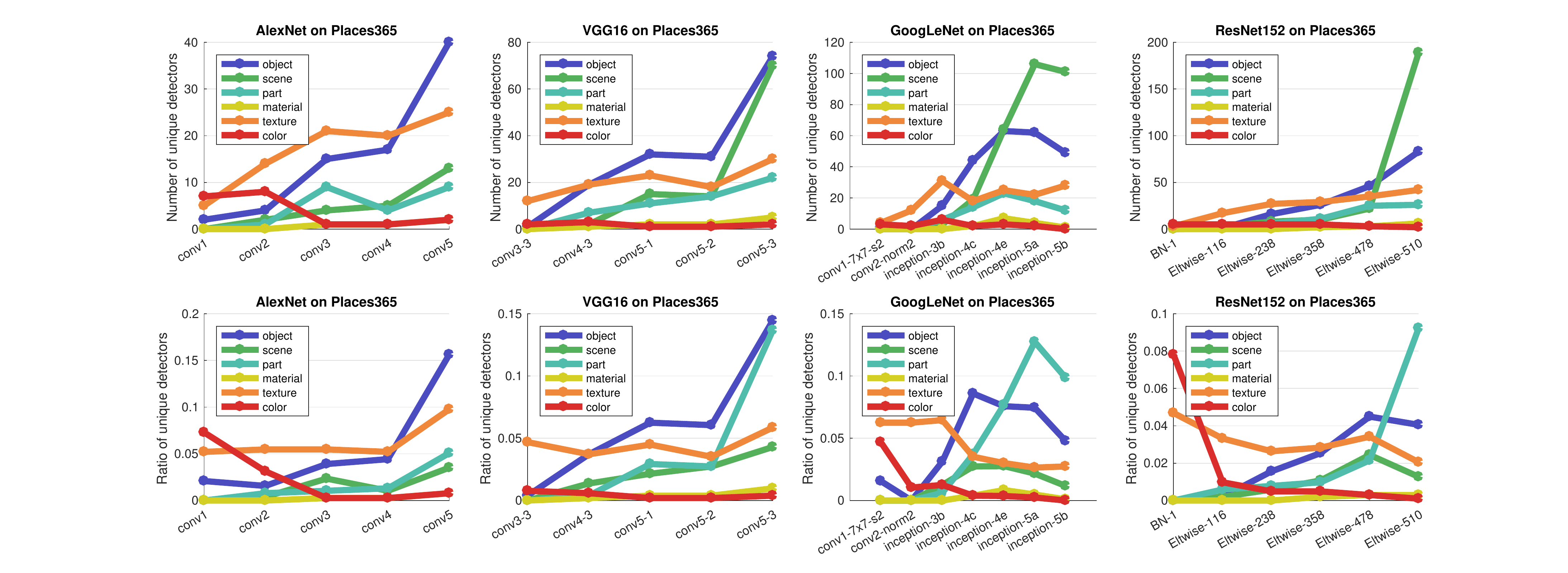}
\vspace{-9mm}
\caption{Comparison of interpretability of the layers for AlexNet, VGG16, GoogLeNet, and ResNet152 trained on Places365. All five conv layers of AlexNet and the selected layers of VGG, GoogLeNet, and ResNet are included. Plot above shows the number of unique detectors and the plot below show the ratio of unique detectors. }\label{fig:alllayer_networks}
\vspace{-5mm}
\end{figure*}

Fig.~\ref{fig:alllayer_networks} shows the interpretable detectors for different layers and network architectures trained on Places365. More object and scene detectors emerge at the higher layers across all architectures, suggesting that representational ability increases with layer depth. %In the plot of the AlexNet-ImageNet and the AlexNet-Places365, we also see that the number of color detectors decreases at higher layers.
%Furthermore, there are more object detectors in the networks trained for scene classification (Places365), compared to the networks trained for object classification (ImageNet).

Because of the compositional structure of the CNN layers, the deeper layers should have higher capacity to represent concepts with larger visual complexity such as objects and scene parts. Our measurements confirm this, and we conclude that higher network depth encourages the emergence of visual concepts with higher semantic complexity.

\subsection{Representations from Self-supervised Learning}
\label{section-selfsupervision}

\begin{figure}
\begin{center}
\includegraphics[width=0.9\linewidth]{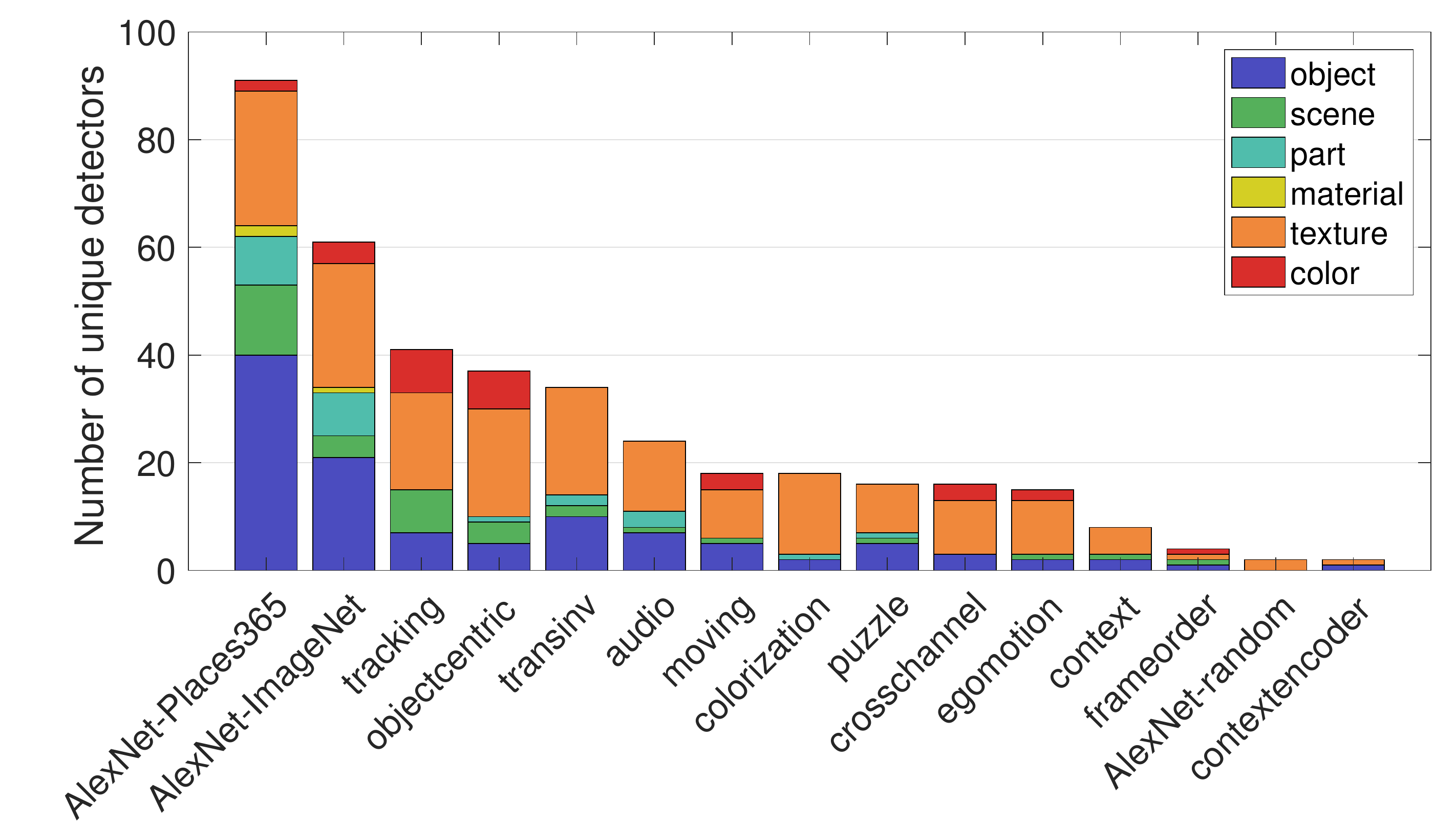}
\end{center}
\vspace{-5mm}
\caption{Semantic detectors emerge across different supervision of the primary training task. All these models use the AlexNet architecture and are tested at \texttt{conv5}.}\label{supervision}
\end{figure}

\begin{figure}
\includegraphics[width=0.48\textwidth]{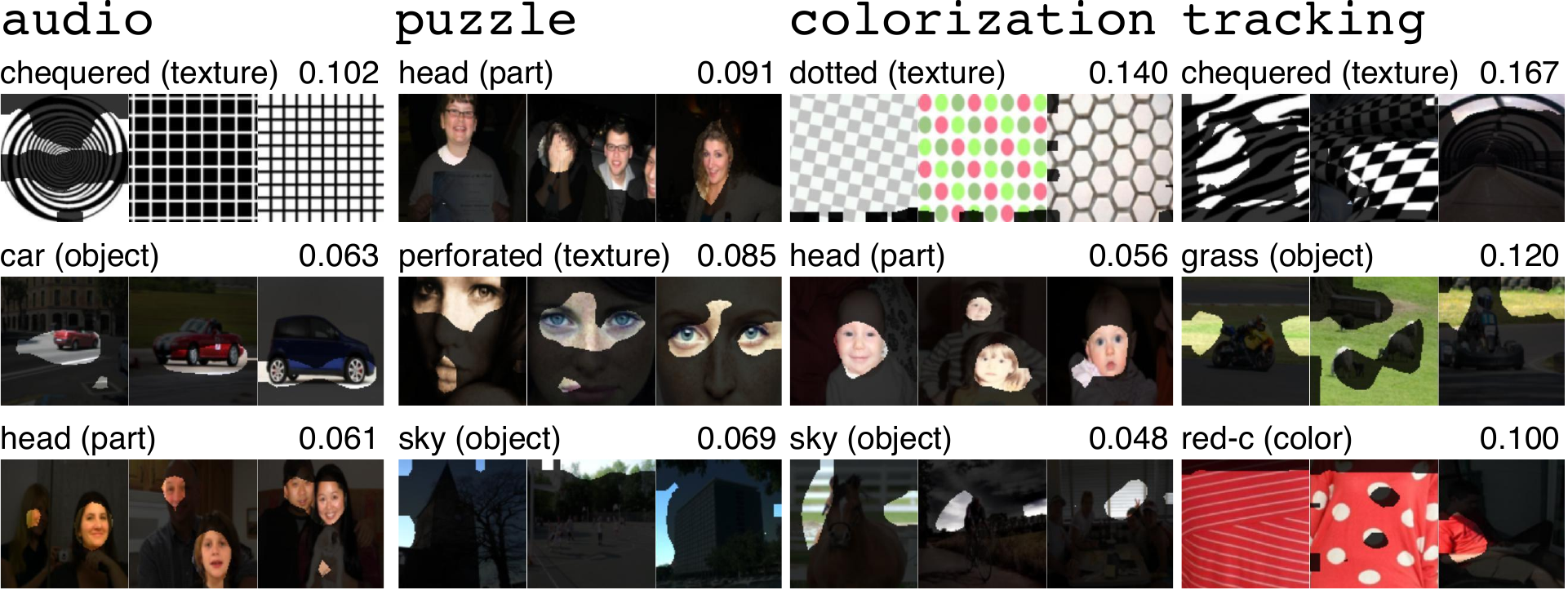}
\vspace{-4mm}
\caption{The top ranked concepts in the three top categories in four self-supervised networks. Some object and part detectors emerge in \texttt{audio}. Detectors for person heads also appear in \texttt{puzzle} and \texttt{colorization}. A variety of texture concepts dominate models with self-supervised training. }\label{fig:selftaught_matter}
\end{figure}

Recently several works have explored a novel paradigm for unsupervised learning of CNNs without using millions of annotated images, namely self-supervised learning. Here, we investigated 12 networks trained for different self-supervised learning tasks: for predicting context (\texttt{context}) \cite{doersch2015unsupervised}, solving puzzles (\texttt{puzzle}) \cite{noroozi2016unsupervised}, predicting ego-motion (\texttt{egomotion})~\cite{jayaraman2015learning}, learning by moving (\texttt{moving})~\cite{agrawal2015learning}, predicting video frame order (\texttt{videoorder})~\cite{misra2016shuffle} or tracking (\texttt{tracking})~\cite{wang2015unsupervised}, detecting object-centric alignment (\texttt{objectcentric}) \cite{gao2016object}, colorizing images (\texttt{colorization})~\cite{zhang2016colorful}, inpainting (\texttt{contextencoder})~\cite{pathak2016context}, predicting cross-channel (\texttt{crosschannel})~\cite{zhang2016splitbrain}, predicting ambient sound from frames (\texttt{audio})~\cite{owens2016ambient}, and tracking invariant patterns in videos (\texttt{transinv})~\cite{wang2017transitive}. The self-supervised models all used AlexNet or an AlexNet-derived architecture, with one exception model \texttt{transinv}~\cite{wang2017transitive}, which uses VGG as the base network.

How do different supervisions affect internal representations? We compared the interpretability resulting from self-supervised learning and supervised learning. We kept the network architecture to AlexNet for each model (one exception is the recent model \texttt{transinv} which uses VGG as the base network). Results are shown in Fig.~\ref{supervision}: training on Places365 creates the largest number of unique detectors. Self-supervised models create many texture detectors but relatively few object detectors; apparently, supervision from a self-taught primary task is much weaker at inferring interpretable concepts than supervised training on a large annotated dataset. The form of self-supervision makes a difference: for example, the colorization model is trained on colorless images, and almost no color detection units emerge. This suggests that emergent units represent concepts required to solve a primary task.

Fig.~\ref{fig:selftaught_matter} shows typical detectors identified in the self-supervised CNN models. For the models \texttt{audio} and \texttt{puzzle}, some part and object detectors emerge. Those detectors may be useful for CNNs to solve primary tasks: the \texttt{audio} model is trained to associate objects with a sound source, so it may be useful to recognize people and cars; while the \texttt{puzzle} model is trained to align the different parts of objects and scenes in an image. For \texttt{colorization} and \texttt{tracking}, recognizing textures might be good enough for the CNN to solve primary tasks such as colorizing a desaturated natural image; thus it is unsurprising that the texture detectors dominate.

\subsection{Representations from Captioning Images}
\label{section-captioning}

\begin{figure}
\begin{center}
\vspace{-4mm}
\includegraphics[width=1\linewidth]{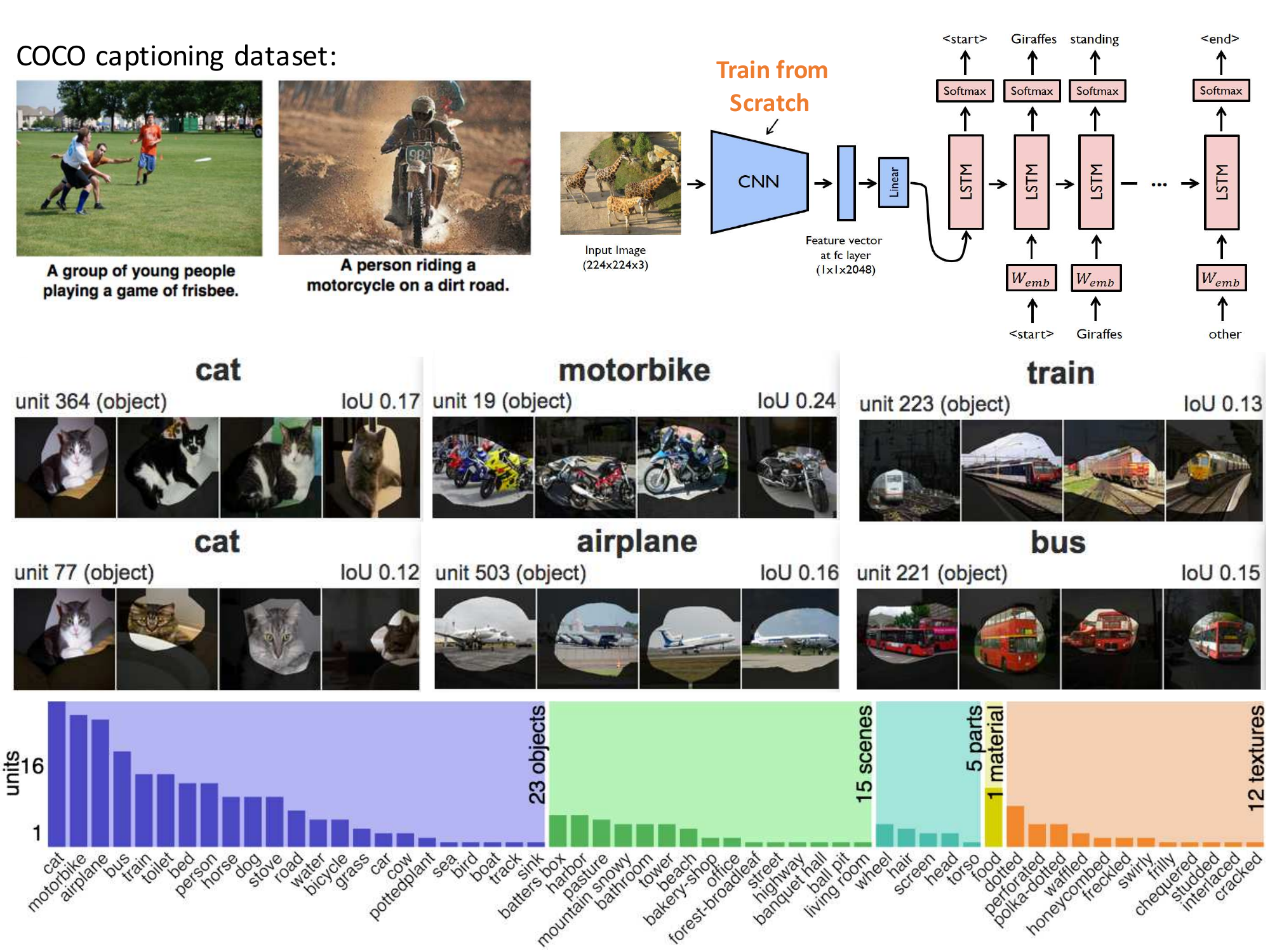}
\end{center}
\vspace{-6mm}
\caption{Example images in the COCO captioning dataset, the CNN+LSTM image captioning model, and the network dissection result. Training ResNet18 from scratch using the supervision from captioning images leads to a lot of emergent object detectors.}\label{coco_captioning}
\end{figure}

To further compare supervised learning and self-supervised learning, we trained a CNN from scratch using the supervision of captioning images, which generates natural language sentence to describe contents. We used the image captioning data from COCO dataset \cite{lin2014microsoft}, with five captions per image. We then trained a CNN plus LSTM as the image captioning model similar to \cite{vinyals2015show}. Features of ResNet18 are used as input to the LSTM for generating captions. The CNN+LSTM architecture and the network dissection results on the last convolutional layer of the ResNet18 are shown in Fig.\ref{coco_captioning}: Many object detectors emerge, suggesting that supervision from natural language captions contains high-level semantics.

\subsection{Training Conditions}
\label{trainingcondition}

\begin{figure}
\begin{center}
\vspace{-4mm}
\includegraphics[width=1\linewidth]{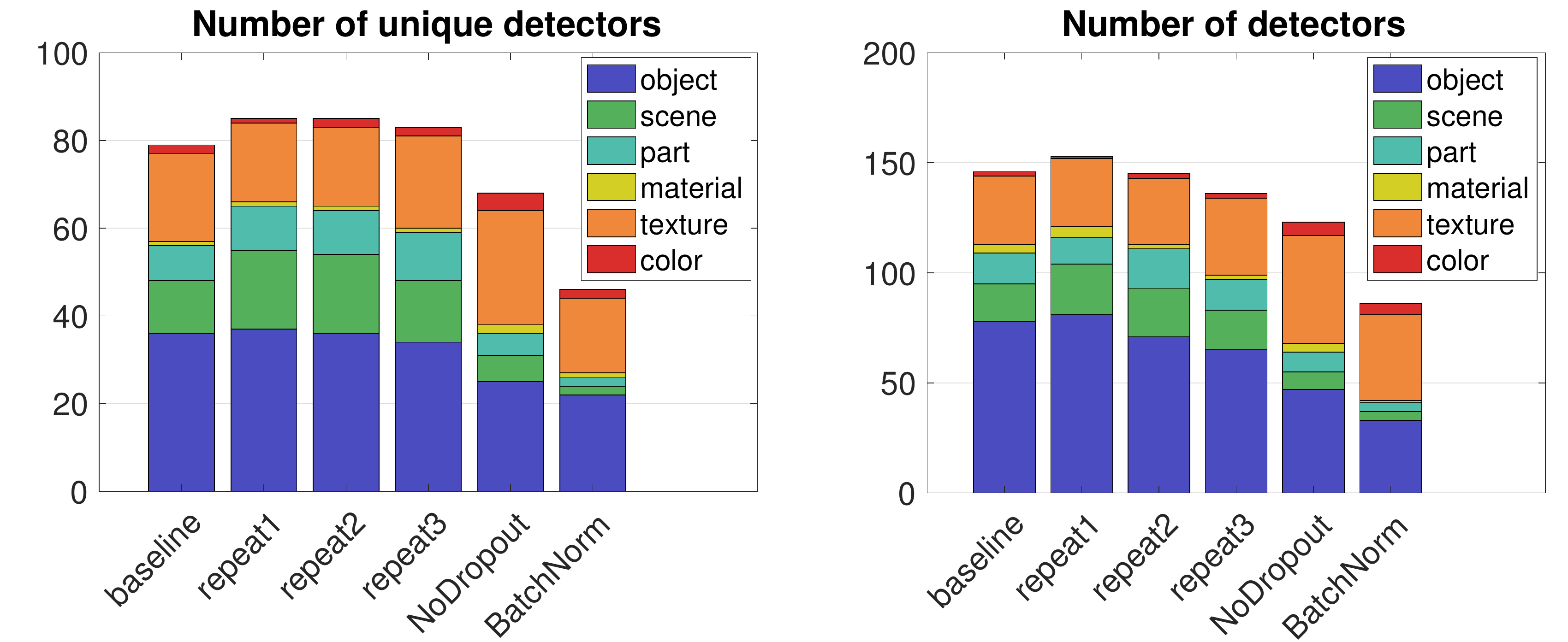}
\end{center}
\vspace{-6mm}
\caption{Effect of regularizations on the interpretability of CNNs. }\label{heuristics}
\vspace{-5mm}
\end{figure}

\begin{figure}
\begin{center}
\vspace{-2mm}
\includegraphics[width=1\linewidth]{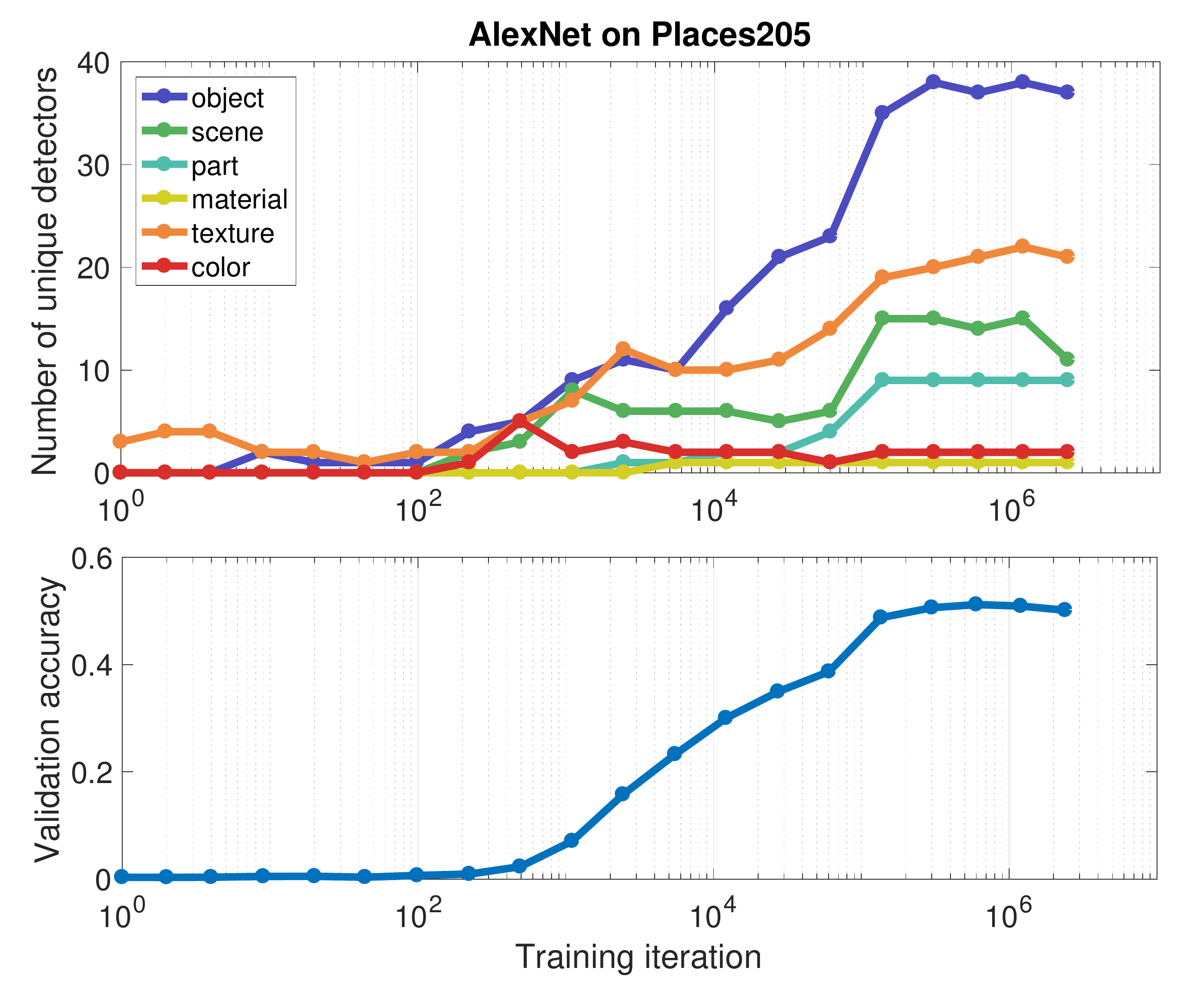}
\end{center}
\vspace{-7mm}
\caption{The evolution of the interpretability of \texttt{conv5} of Places205-AlexNet over 3,000,000 training iterations. The accuracy on the validation at each iteration is also plotted. The baseline model is trained to 300,000 iterations (marked at the red line).}\label{iterations}
\end{figure}

\begin{figure*}
\includegraphics[width=1\linewidth]{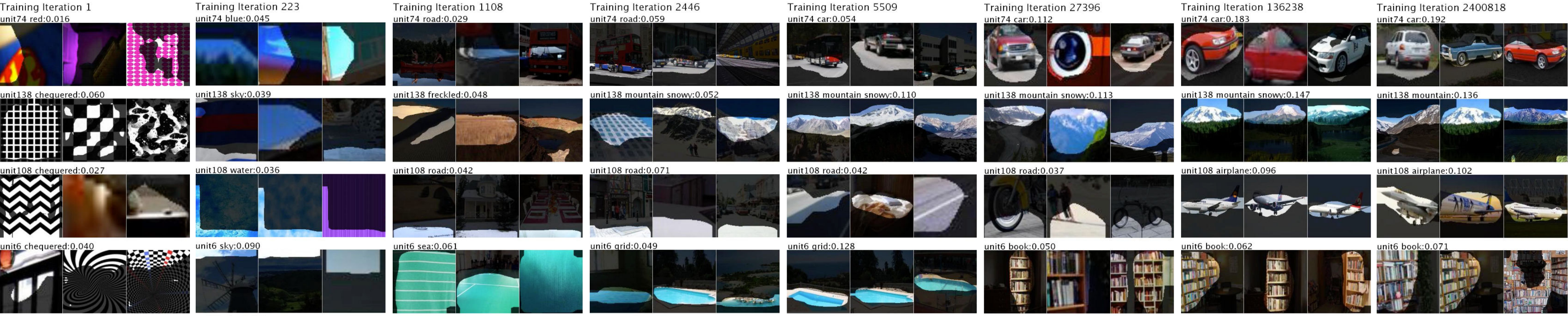}
\vspace{-8mm}
\caption{The interpretations of units change over iterations. Each row shows the interpretation of one unit.}\label{fig:unitmutation_iteration}
\vspace{-5mm}
\end{figure*}

The number of training iterations, dropout \cite{srivastava2014dropout}, batch normalization \cite{ioffe2015batch}, and random initialization \cite{li2015convergent}, are known to affect the representation learned by neural networks. To analyze the effect of training conditions on interpretability, we took Places205-AlexNet as the baseline model and prepared several variants of it, all using the same AlexNet architecture.
For the variants \textit{Repeat1},  \textit{Repeat2} and \textit{Repeat3}, we randomly initialized the weights and trained them with the same number of iterations. For the variant \textit{NoDropout}, we removed the dropout in the FC layers of the baseline model. For the variant \textit{BatchNorm}, we applied batch normalization at each convolutional layer of the baseline model. \textit{Repeat1}, \textit{Repeat2}, \textit{Repeat3} all have nearly the same top-1 accuracy 50.0\% on the validation set. The variant without dropout has top-1 accuracy 49.2\%. The variant with batch norm has top-1 accuracy 50.5\%.

Fig.~\ref{heuristics} shows the results: 1) Comparing different random initializations, the models converge to similar levels of interpretability, both in terms of unique detector number and total detector number; this matches observations of convergent learning discussed in \cite{li2015convergent}. 2) For the network without dropout, more texture detectors but fewer object detectors, emerge. 3) Batch normalization seems to decrease interpretability significantly.

The batch normalization result serves as a caution that discriminative power is not the only property of a representation that should be measured.  Our intuition here is that the batch normalization `whitens' the activation at each layer, which smooths out scaling issues and allows a network to easily rotate axes of intermediate representations during training.  While whitening apparently speeds training, it may also have an effect similar to random rotations analyzed in Sec.~\ref{section-rotation} which destroy interpretability. As discussed in Sec.~\ref{section-rotation}, however, interpretability is neither a prerequisite nor an obstacle to discriminative power.  Finding ways to capture the benefits of batch normalization without destroying interpretability is an important area for future work.

Fig.~\ref{iterations} plots the interpretability of snapshots of the baseline model at different training iterations along with the accuracy on the validation set. We can see that object detectors and part detectors begin emerging at about 10,000 iterations (each iteration processes a batch of 256 images). We do not find evidence of transitions across different concept categories during training.  For example, units in \texttt{conv5} do not turn into texture or material detectors before becoming object or part detectors. In Fig.~\ref{fig:unitmutation_iteration}, we keep track of six units over different training iteration. We observe that some units start converging to the semantic concept at early stage. For example, unit138 starts detecting mountain snowy as early as at iteration 2446. We also observe that units evolve over time: unit74 and unit108 detect road first before they start detecting car and airplane respectively.

\begin{figure}
\includegraphics[width=1\linewidth]{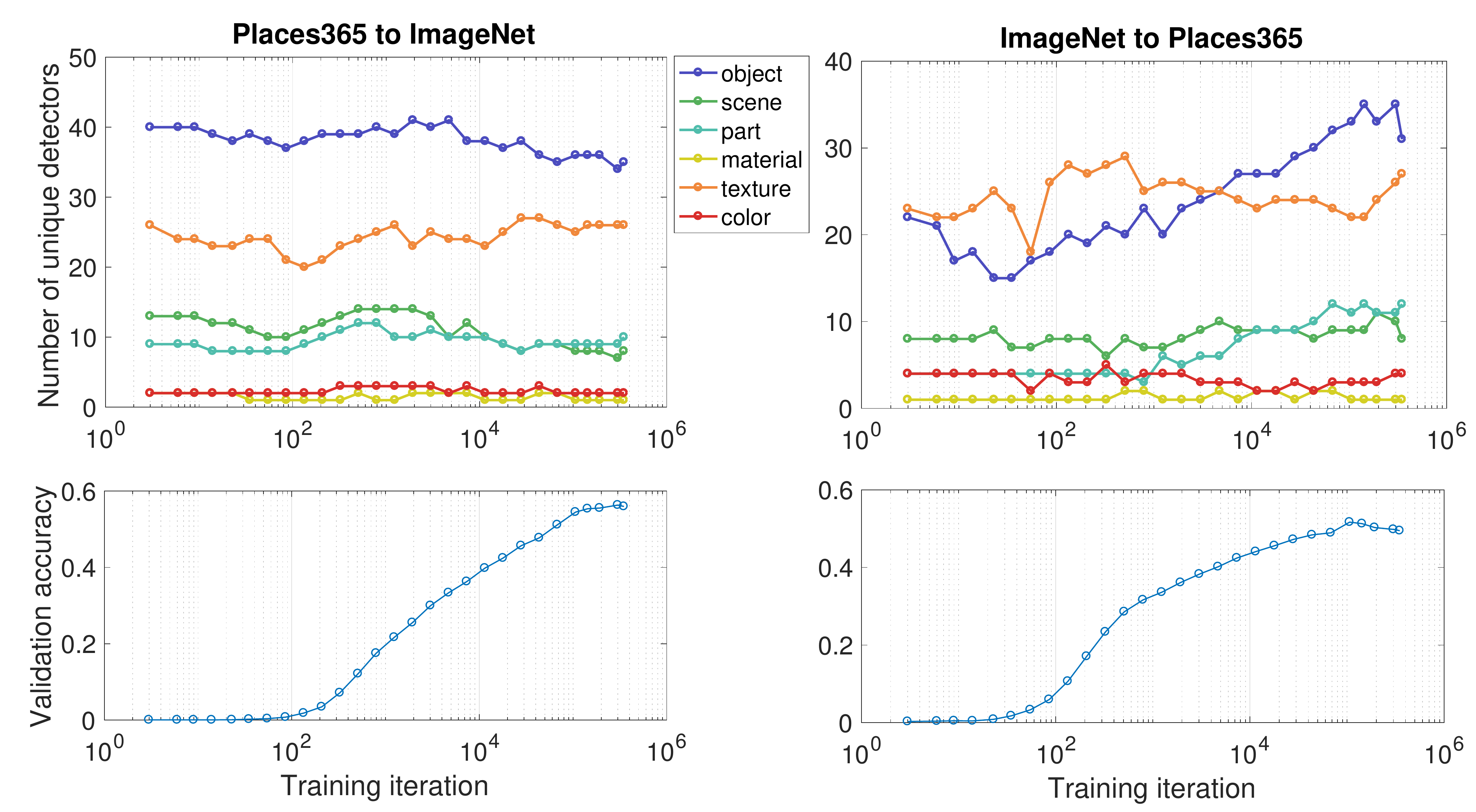}
\vspace{-8mm}
\caption{a) Fine-tune AlexNet from ImageNet to Places365. b) Fine-tune AlexNet from Places365 to ImageNet.}\label{fig:finetune}
\end{figure}
 
\begin{figure}
\includegraphics[width=1\linewidth]{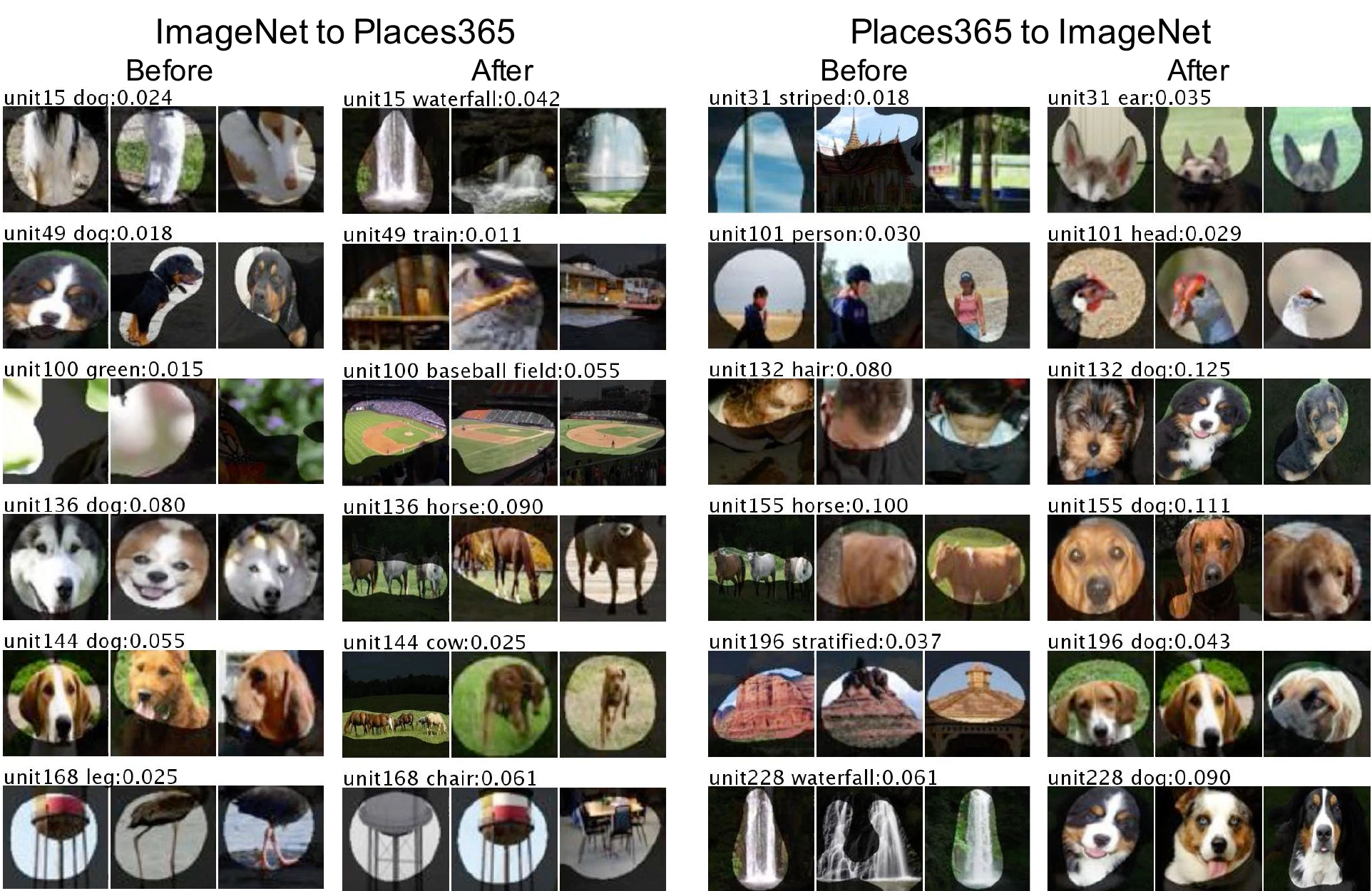}
\vspace{-7mm}
\caption{Units evolve from a) the network fine-tuned from ImageNet to Places365 and b) the network fine-tuned from Places365 to ImageNet. Six units are shown with their semantics at the beginning of the fine-tuning and at the end of the fine-tuning. }\label{fig:unitmutation_finetune}
\end{figure}

\subsection{Transfer Learning between Places and ImageNet}
\label{section-transfer-learning}
Fine-tuning a pre-trained network to a target domain is commonly used in transfer learning. The deep features from the pre-trained network show good generalization across different domains. The pre-trained network also makes the training converge faster and results in better accuracy, especially if there is not enough training data for the target domain. Here we analyze how unit interpretation evolv during transfer learning.

To see how individual units evolve across domains, we run two experiments: fine-tuning Places-AlexNet to ImageNet and fine-tuning ImageNet-AlexNet to Places. The interpretability results of the model checkpoints at different fine-tuning iteration are plotted in Fig.~\ref{fig:finetune}. The training indeed converges faster compared to the network trained from scratch on Places in Fig.~\ref{iterations}. The interpretations of the units also change over fine-tuning. For example, the number of unique object detectors first drop then keep increasing for the network trained on ImageNet being fine-tuned to Places365, while it is slowly dropping for the network trained on Places being fine-tuned to ImageNet.

Fig.~\ref{fig:unitmutation_finetune} shows some examples of the individual unit evolution happening in the networks trained from ImageNet to Places365 and from Places365 to ImageNet, at the beginning and at the end of fine-tuning. In the ImageNet to Places365 network, unit15 which detects white dogs initially, evolves to detect waterfall; unit136 and unit144 which detect dogs first, evolve to detect horse and cow respectively (note a lot of scene categories in Places like pasture and corral contain these animals). In the Places365 to ImageNet network, several units evolve to be dog detectors, given ImageNet distribution of categories. While units evolve to detect different concepts, the before and after- concepts often share low-level image similarity such as colors and textures. 

\begin{figure}
\includegraphics[width=1\linewidth]{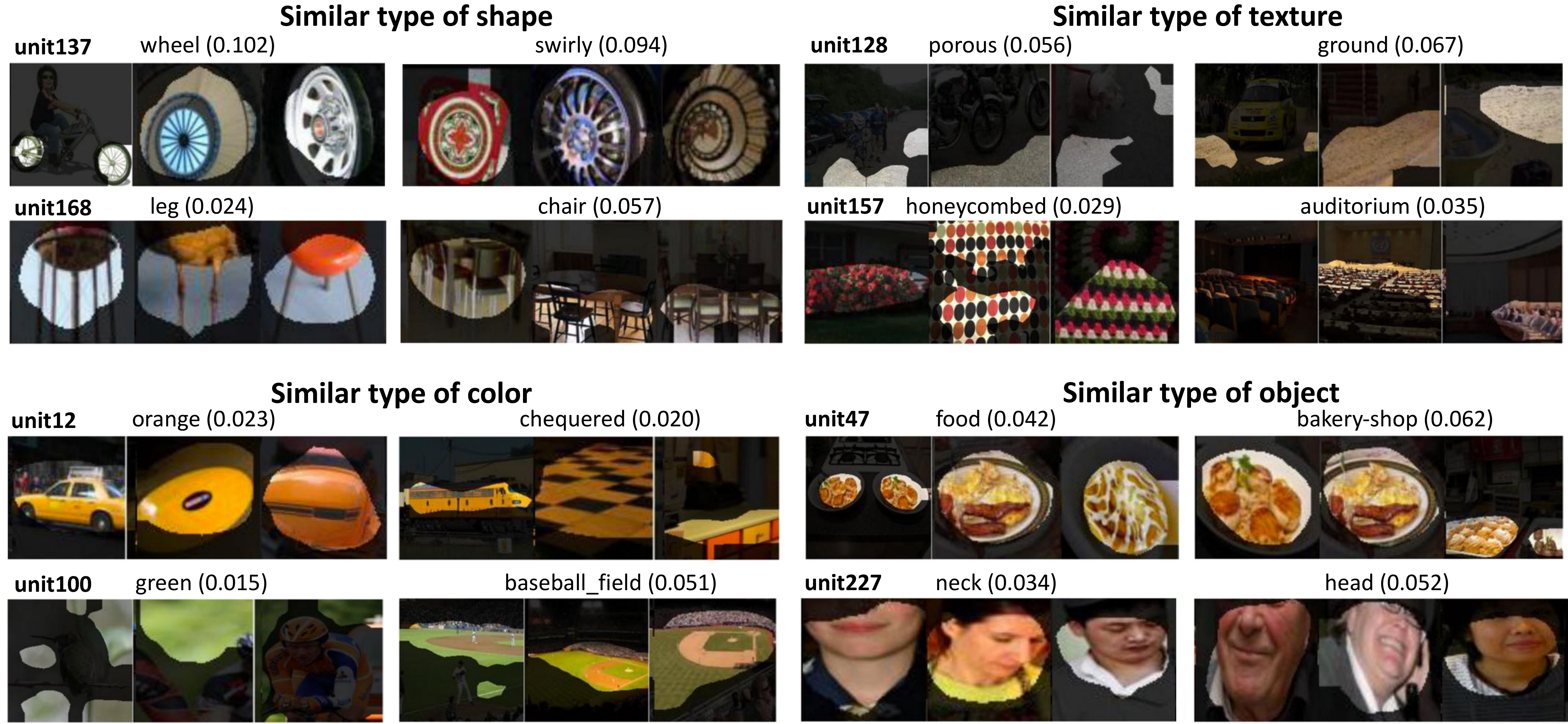}
\vspace{-7mm}
\caption{Examples from four types of unit evolutions. Types are defined based on the concept similarity.}\label{fig:finetune_type}
\end{figure}

\begin{figure}
\includegraphics[width=1\linewidth]{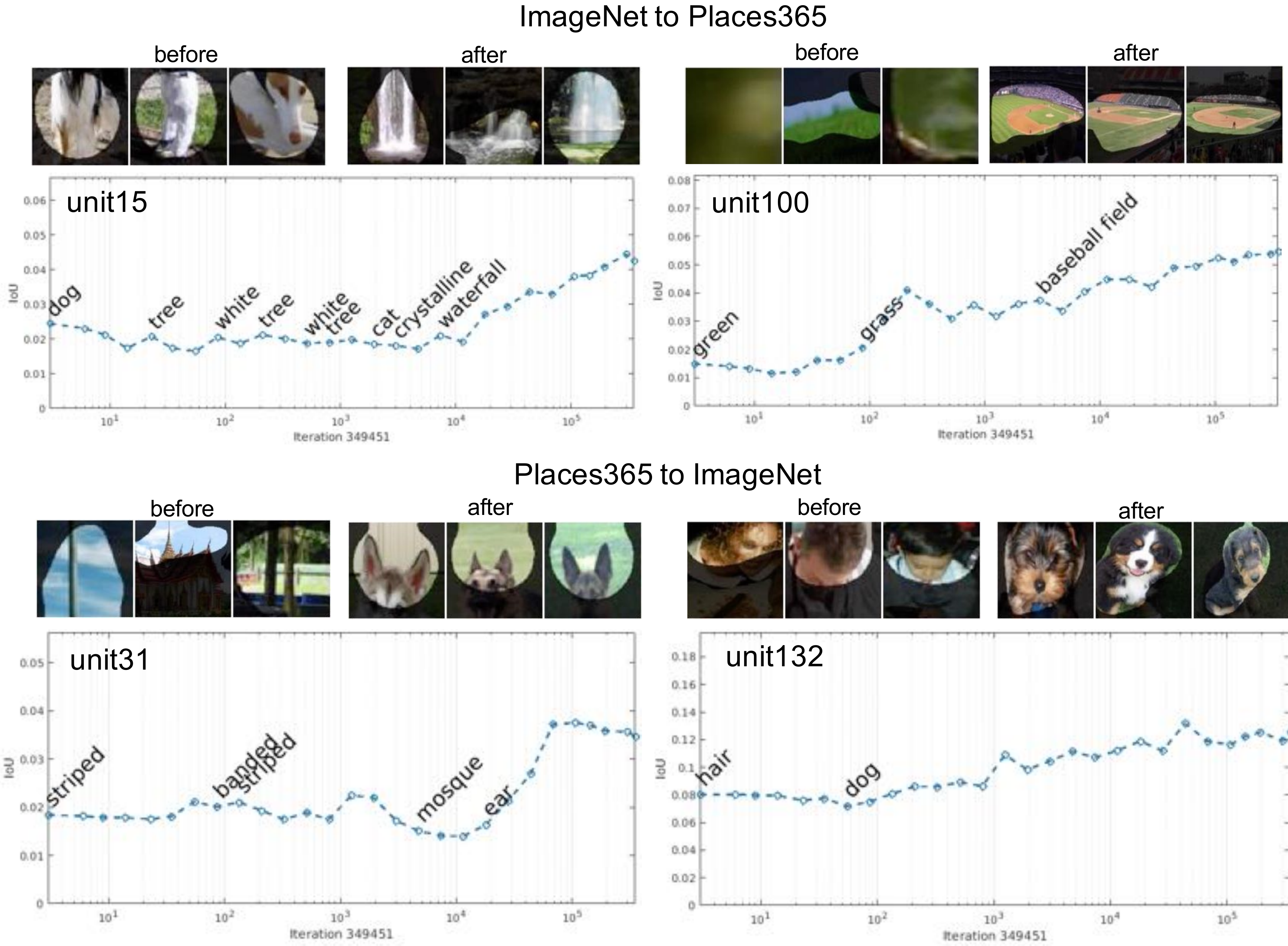}
\vspace{-7mm}
\caption{The history of one unit evolution during the fine-tuning from ImageNet to Places365 (top) and Places365 to ImageNet (low).}\label{fig:finetune_history}
\end{figure}

The fine-tuned model achieves almost the same classification accuracy as the train-from-scratch model, but the training converges faster due to the feature reuse. For the ImageNet to Places network, 139 out of 256 units (54.4\%) at conv5 layer keep the same concepts during the finetuning, while for the network fine-tuned from Places to ImageNet, 135 out of 256 units (52.7\%) at conv5 stay have the same concepts. We further categorized the unit evolution into five types based on the similarity between the concepts before and after fine-tuning. Out of the 117 units which evolved in the network fine-tuned from Imagenet to Places, 47 units keep a similar type of shape, 31 units have a similar texture, 18 units have similar colors, 13 units have a similar type of object, and 8 units do not have a clear pattern of similarities (see Fig.\ref{fig:finetune_type}). Fig.~\ref{fig:finetune_history} illustrates the evolution history for two units of each model. Units seem to switch their top ranked label times before converging to a concept: unit15 in the fine-tuning of ImageNet to Places365 flipped to white, crystalline, before stabilizing to a waterfall concept. Other units are switching faster: unit132 in the fine-tuning of Places365 to ImageNet goes from hair to dog at an early stage of fine-tuning.

\begin{figure}
\includegraphics[width=1\linewidth]{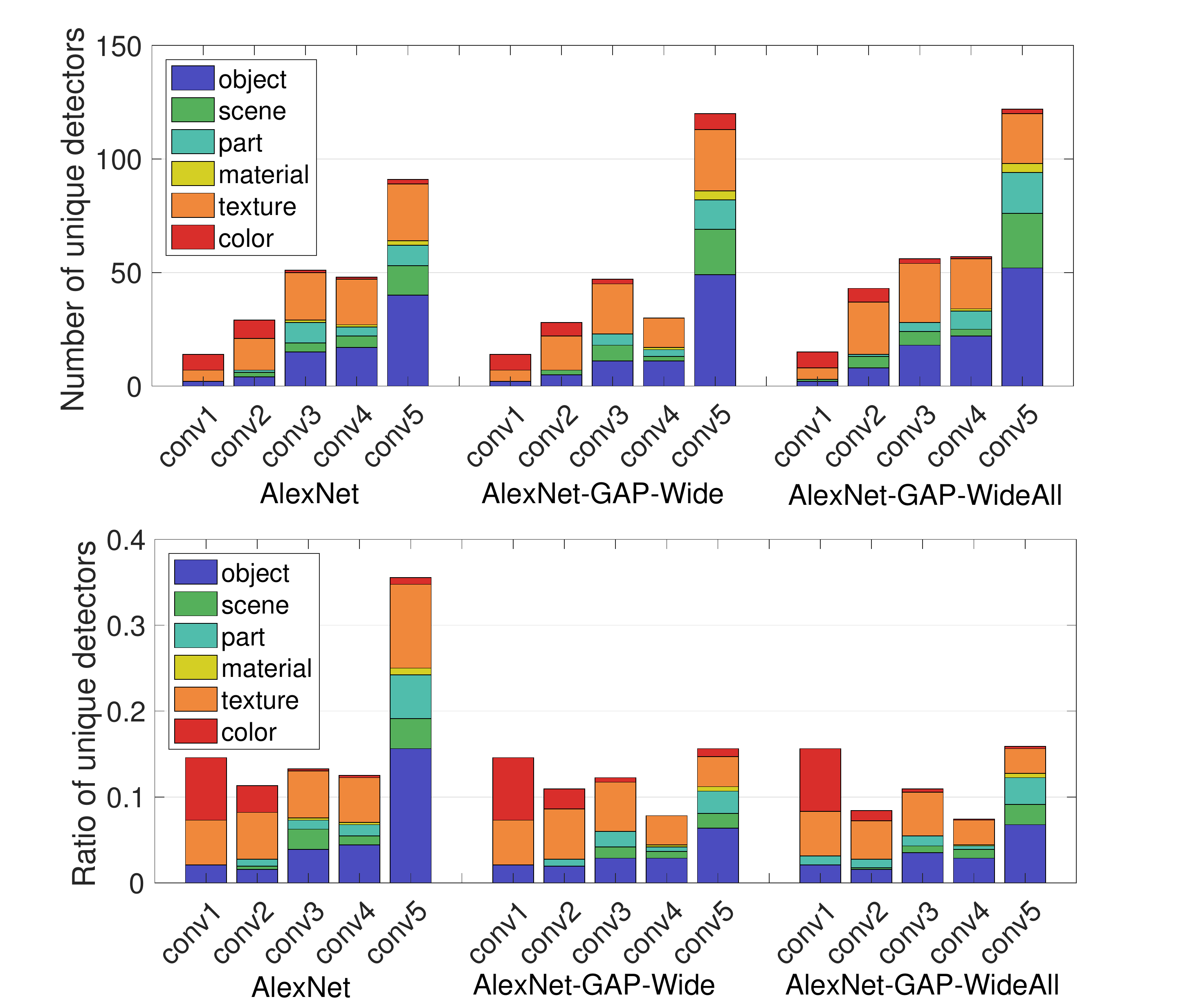}
\vspace{-8mm}
\caption{Comparison of the standard AlexNet, AlexNet-GAP-Wide, and AlexNet-GAP-WideAll. Widening the layer brings the emergence of more detectors. Networks are trained on Places365. Plot above shows the number of unique detectors, plot below shows the ratio of unique detectors.}\label{fig:width_matter}
\end{figure}

\subsection{Layer Width vs. Interpretability}
\label{section-layerwidth}
From AlexNet to ResNet, CNNs have grown deeper in the quest for higher classification accuracy. Depth is important for high discrimination ability, and as shown in Sec.~\ref{section-comparing-architecture}, interpretability increases with depth. However, the role of the width of layers (the number of units per layer) has been less explored. One reason is that increasing the number of convolutional units in a layer significantly increases computational cost while yielding only marginal classification accuracy improvements. Nevertheless, some recent work \cite{zagoruyko2016wide} suggests that a carefully designed wide residual network can achieve classification accuracy superior to the commonly used thin and deep counterparts.

To test how width affects emergence of interpretable detectors, we removed the FC layers of AlexNet, then tripled the number of units at the \texttt{conv5}, \textit{i.e.}, from 256 to 768 units, as \textit{AlexNet-GAP-Wide}. We further tripled the number of units for all the previous conv layers except conv1 for the standard AlexNet, as \textit{AlexNet-GAP-WideAll}. Finally we put a global average pooling layer after \texttt{conv5} and fully connected the pooled 768-feature activations to the final class prediction. After training on Places365, the AlexNet-GAP-Wide and the AlexNet-GAP-WideAll have similar classification accuracy on the validation set as the standard AlexNet ($\sim0.5\%$ top1 accuracy lower and higher): however many more emergent unique concept detectors at \texttt{conv5} are found for AlexNet-GAP-Wide and all the conv layers for AlexNet-GAL-WideAll (see Fig.~\ref{fig:width_matter}). Increasing the number of units to 1024 and 2048 at \texttt{conv5}, did not significantly increase the unique concepts. This may indicate either a limit on the capacity of AlexNet to separate explanatory factors, or a limit on the number of disentangled concepts that are helpful to solve the primary task of scene classification.

\subsection{Discrimination vs. Interpretability}
\label{section-discrimination}
\begin{figure}
\begin{center}
\includegraphics[width=1\linewidth]{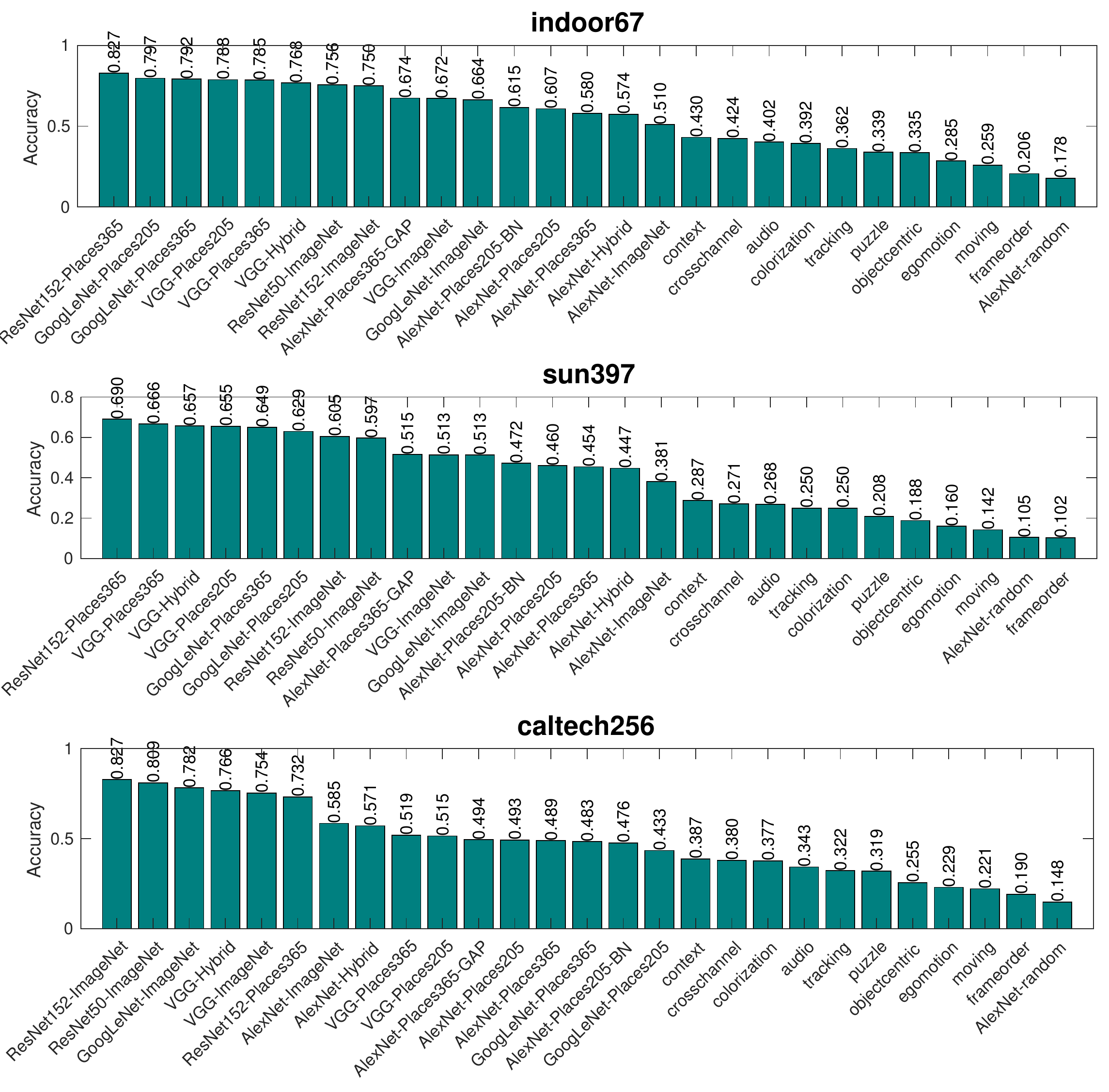}
\end{center}
\vspace{-5mm}
\caption{The classification accuracy of deep features on the three image datasets.}\label{classification_accuracy}
\end{figure}

Activations from the higher layers of pre-trained CNNs are often used as generic visual features (noted as deep features), generalizing well to other image datasets \cite{zhou2014learning,razavian2014cnn}. It is interesting to bridge the notion of generic visual features with their interpretability. Here we first benchmarked the deep features from several networks on several image classification datasets for their discriminative power. For each network, we fed in the images and extracted the activation at the last convolutional layer as the visual feature. Then we trained a linear SVM with $C=0.001$ on the train split and evaluated the performance on the test split. We computed the classification accuracy averaged across classes, see Fig.~\ref{classification_accuracy}. We include indoor67 \cite{quattoni2009recognizing}, sun397 \cite{xiao2010sun} and caltech256 \cite{griffin2007caltech}. 
The deep features from supervised trained networks perform much better than the ones from the self-supervised trained networks. Networks trained on Places have better features for scene-centric datasets (sun397 and indoor67), while networks trained on ImageNet have better features for object-centric datasets (caltech256).

% \begin{table}\caption{Statistics of the datasets.}
% \label{stat_dataset}
% \centering
% \tiny
% \begin{tabular}{| l | c | c | c | c | c | c |c |}
%   \hline                       
% Dataset & event8 & action40  & indoor67 & sun397  & caltech101 & caltech256 \\
% \hline
% Image No. & 1,574 & 9,532 & 15,613 & 108,770 & 8,677 & 29,780 \\
% Class No. & 8 & 40 & 67 & 397 & 101 & 256 \\
% Sample/class & 70 & 100 & 100 & 50 & 30 & 30 \\
% Train No. & 560 & 4,000 & 6,700 & 19,850 & 3,030 & 7,680 \\
% Test No.  & 1,014 & 5,532 & 8,913 & 88,920 & 5647 & 22,100 \\
% \hline
% \end{tabular}
% \end{table}

\begin{figure*}
\begin{center}
\includegraphics[width=0.9\textwidth]{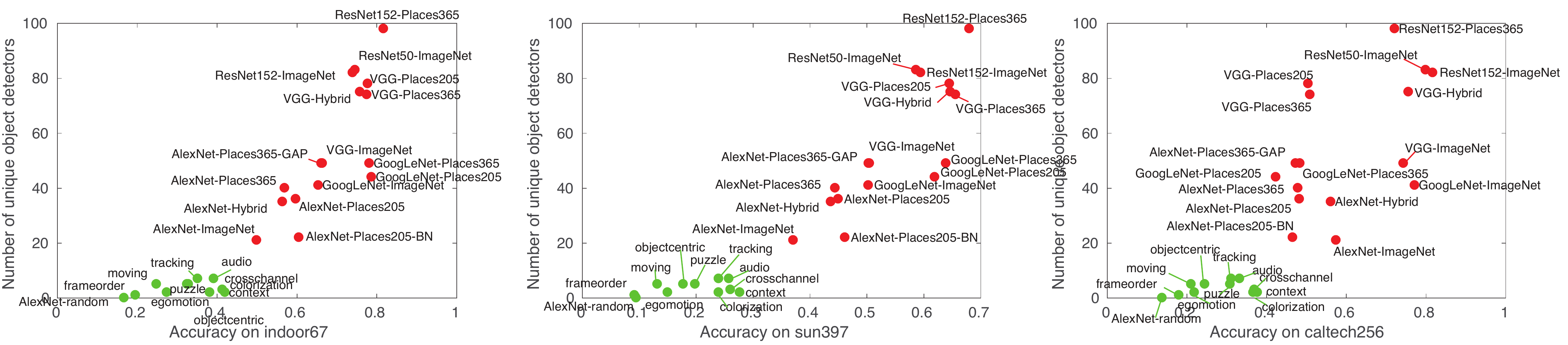}
\end{center}
\vspace{-6mm}
\caption{The number of unique object detectors in the last convolutional layer compared to each representation’s classification accuracy on three datasets. Supervised (in red) and unsupervised (in green) representations clearly form two clusters.}
\label{semanticsVSclassification}
\vspace{-3mm}
\end{figure*}

\begin{figure*}[!b]
\begin{center}
\includegraphics[width=1\linewidth]{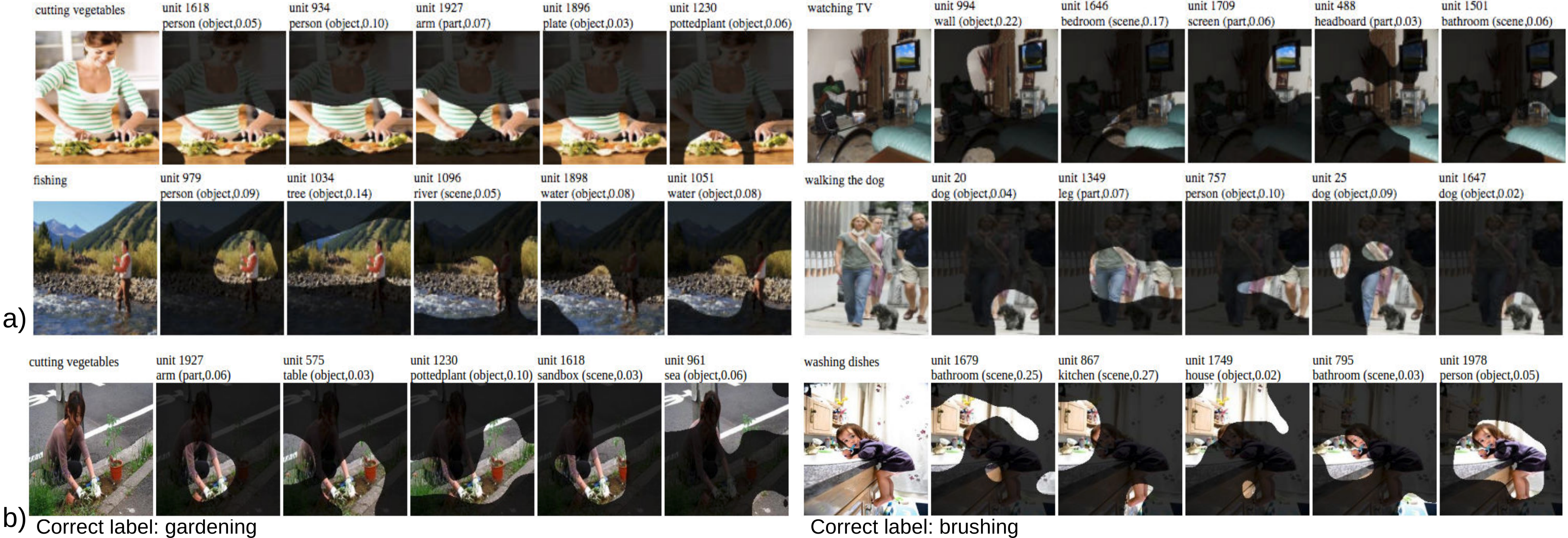}
\end{center}
\vspace{-5mm}
\caption{Segmenting images using top activated units weighted by the class label from ResNet152-Places365 deep feature. a) the correctly predicted samples. b) the incorrectly predicted samples.}\label{fig:unit_segmentation}
\end{figure*}

Fig.~\ref{semanticsVSclassification} plots the number of the unique object detectors for each representation over that representation's classification accuracy on three selected datasets. There is positive correlation between them suggesting that the supervision tasks that encourage the emergence of more concept detectors may also improve the discrimination ability of deep features. Interestingly, on some of the object centric dataset, the best discriminative representation is the representation from ResNet152-ImageNet, which has fewer unique object detectors compared to the ResNet152-Places365. We hypothesize that the accuracy on a representation when applied to a task is dependent not only on the number of concept detectors in the representation, but on how well the concept detectors captures the characteristics of the hidden factors in the transferred dataset.

% \begin{figure}
% \begin{center}
% \includegraphics[width=.95\linewidth]{plot/camera_semanticsVSclassification.pdf}
% \end{center}
% \caption{The number of unique object detectors in the last convolutional layer compared to each representation’s classification accuracy on the action40 dataset. Supervised and unsupervised representations clearly form two clusters.}\label{discrimination}
% \end{figure}

\subsection{Explaining the Predictions for the Deep Features}
\label{section-explaination}
After we interpret the units inside the deep visual representation, we show that the unit activation along with the interpreted label can be used to explain the prediction given by the deep features. Previous work \cite{zhou2016cvpr} uses the weighted sum of the unit activation maps to highlight which image regions are most informative to the prediction, here we further decouple at individual unit level to segment the informative image regions.

We use the individual units identified as concept detectors to build an explanation of the individual image prediction given by a classifier. The procedure is as follows: Given any image, let the unit activation of the deep feature (for ResNet the \texttt{GAP} activation) be $[x_1, x_2, ..., x_N]$, where each $x_{n}$ represents the value summed up from the activation map of unit $n$. Let the top prediction's SVM response be $s = \sum_{n}w_{n}x_{n}$, where $[w_1, w_2, ..., w_N]$ is the SVM's learned weight. We get the top ranked units in Figure \ref{fig:unit_segmentation} by ranking $[w_{1}x_1, w_{2}x_2, ..., w_{N}x_N]$, which are the unit activations weighted by the SVM weight for the top predicted class. Then we simply upsample the activation map of the top ranked unit to segment the image.  The threshold used for segmentation is the top 0.2 activation of the unit based on the feature map of the single instance.

Image segmentations using individual unit activation on action40 \cite{yao2011human} dataset are plotted in Fig.~\ref{fig:unit_segmentation}a. The unit segmentation explain the prediction explicitly. For example, the prediction for the first image is \textit{Gardening}, and the explanatory units detect person, arm, plate, pottedplant. The prediction for the second image is \textit{Fishing}, the explanatory units detect person, tree, river, water. We also plot some incorrectly predicted samples in Figure \ref{fig:unit_segmentation}b. The segmentation gives the intuition as to why the classifier made mistakes. For example, for the first image the classifier predicts \textit{cutting vegetables} rather than the true label \textit{gardening}, because the second unit incorrectly mistakes the ground as table.

\section{Discussion}

We discuss the threshold $\tau$ and the potential biases in the interpretation given by our approach below.

\begin{figure}
\centering
\includegraphics[width=\columnwidth]{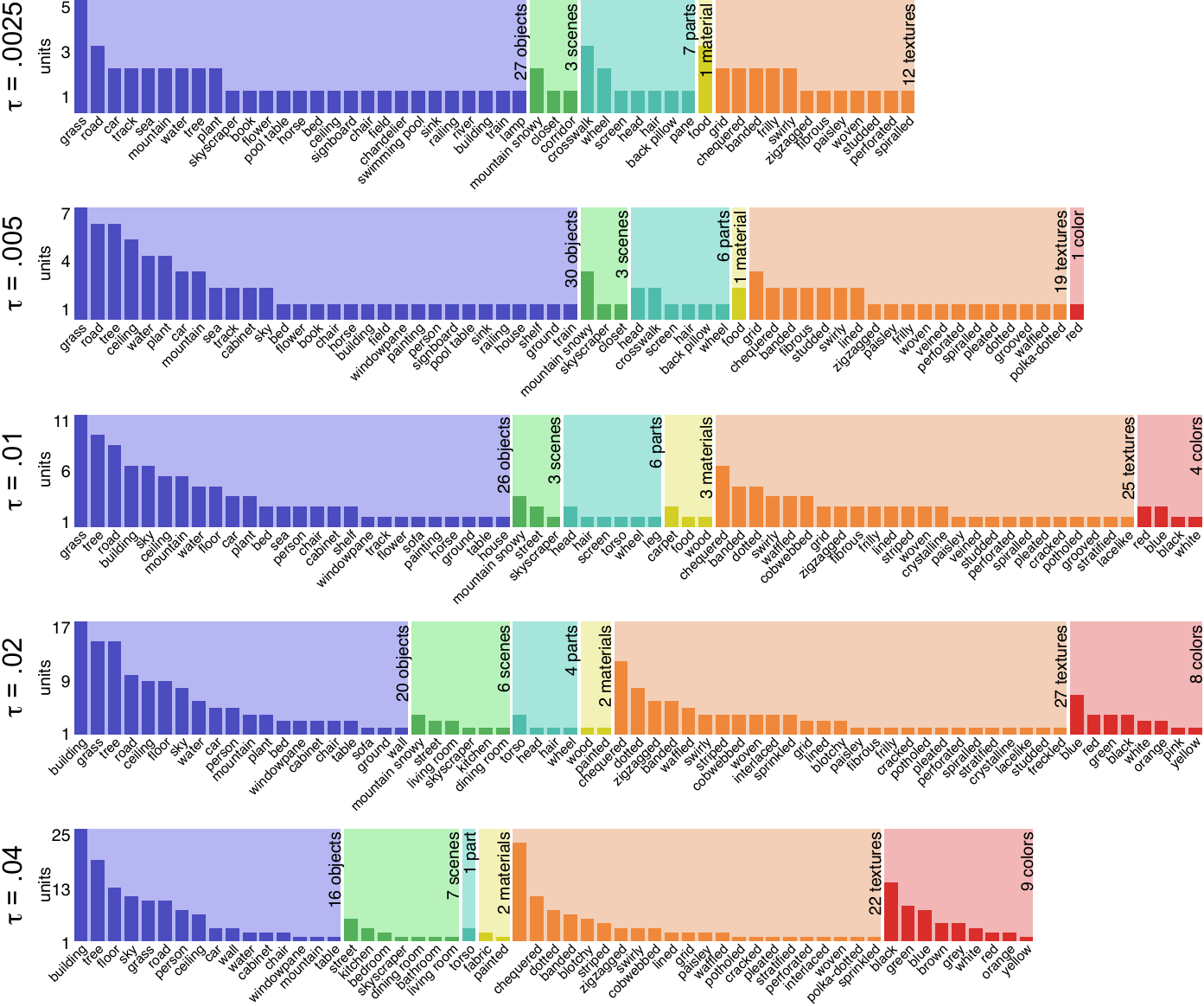}
\caption{Labels that appear in Alexnet-conv5 on Places205 as $\tau$ is varied from 0.0025 to 0.04.  At wider thresholds, more units are assigned to labels for generic concepts such as colors and textures.}
\label{fig:varying-tau}
\end{figure}

\begin{figure}
\centering
\subfloat[Unit 46 ``painting'' top images]{
\includegraphics[width=0.48\columnwidth]{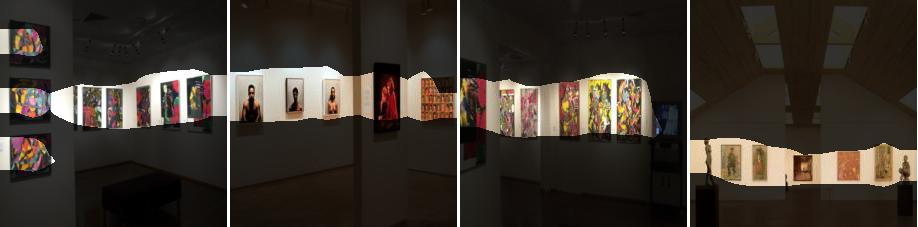}}\hfill
\subfloat[Unit 185 ``horse'' top images]{
\includegraphics[width=0.48\columnwidth]{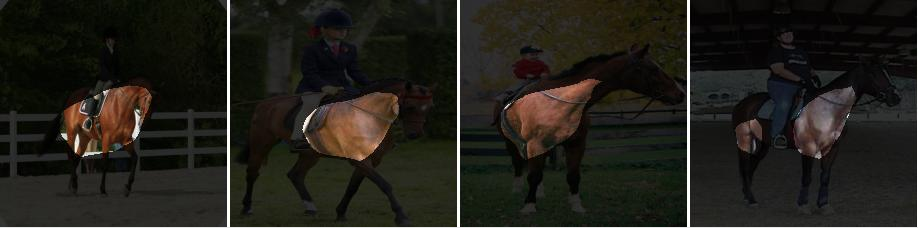}} \\
\subfloat[IoU of labels matching unit 46 at different $\tau$.]{
\includegraphics[width=0.48\columnwidth,trim={8mm 0 14mm 0}]{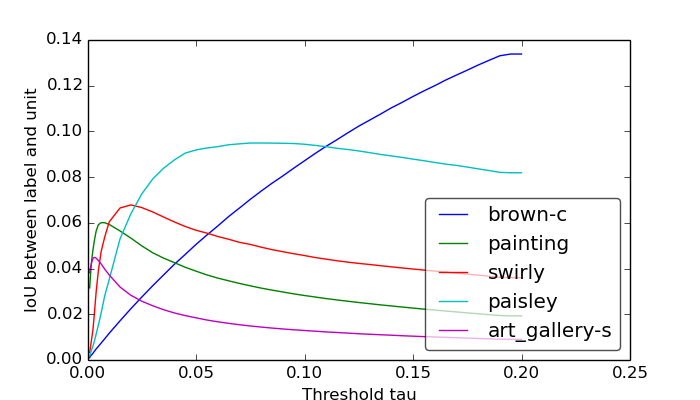}}\hfill
\subfloat[IoU of lables matching unit 185 at different $\tau$.]{
\includegraphics[width=0.48\columnwidth,trim={8mm 0 14mm 0}]{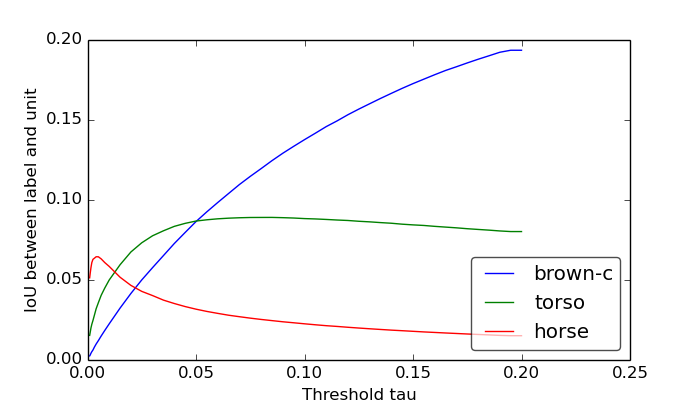}} \\
\caption{Typical relationships between $\tau$ and IoU for different labels.  In (c) and (d), IoU is shown on the $y$ axis and $\tau$ is on the x axis, and every concept in Broden which maximizes IoU for some $\tau$ is shown.  For loose thresholds, the same general concept ``brown color'' maximizes IoU for both units even though the units have remarkable distinctive selectivity at tighter thresholds.}
\label{fig:tau-to-iqr}
\end{figure}

\textbf{Influence of the threshold $\tau$}. Our choice of a tight threshold $\tau$ is done to reveal information about fine-grained concept selectivity of individual units.  The effect of choosing tighter and looser $\tau$ on the interpretation of units across a whole representation is shown in Fig~\ref{fig:varying-tau}.  A $\tau$ smaller than $0.005$ identifies fewer objects because some objects will be missed by the small threshold.  On the other hand, a larger $\tau$, or using no threshold at all, associates units with general concepts such as colors, textures, and large regions, rather than capturing the sensitivity of units on more specific concepts.  Fig.~\ref{fig:tau-to-iqr} shows the effect of varying $\tau$ on specific units' IoU.  Although the two units are sensitive to paintings and horses, respectively, they are also both generally sensitive to the color brown when considered at a larger $\tau$.  The tight $\tau=0.005$ reveals the sensitivity of the units to fine-grained concepts.

\textbf{Potential biases in the interpretations}. Several potential biases might occur to our method as follows: 1) Our method will not identify units that detect concepts that do not appear in the Broden dataset, including some difficult-to-name concepts such as `the corner of a room'; 2) Some units might detect a very fine-grained concept, such as a wooden stool chair leg, which are more specific than concepts in Broden, thus yielding a low IoU on the `chair' category. Such units might not be counted as a concept detector. 3) Our method measures the degree of alignment between individual unit activations and a visual concept, so it will not identify a group of units that might jointly represent one concept; 4) Units might not be centered within their receptive fields so that the upsampled activation maps may be misaligned by a few pixels. 5) The "number of unique detectors" metric might favor large networks in comparing their network interpretability.

\section{Conclusion}

Network Dissection translates qualitative visualizations of representation units into quantitative interpretations and measurements of interpretability. Here we show that the units of a deep representation are significantly more interpretable than expected for a basis of the representation space.  We investigate the interpretability of deep visual representations resulting from different architectures, training supervisions, and training conditions. We also show that interpretability of deep visual representations is relevant to the power of the representation as a generalizable visual feature. We conclude that interpretability is an important property of deep neural networks that provides new insights into their hierarchical structure. Our work motivates future work towards building more interpretable and explainable AI systems.

%This paper proposed a general framework, network dissection, for quantifying interpretability of CNNs. We applied network dissection to measure whether interpretability is an axis-independent phenomenon, and we found that it is not. This is consistent with the hypothesis that interpretable units indicate a partially disentangled representation. We applied network dissection to investigate the effects on interpretability of state-of-the art CNN training techniques. We have confirmed that representations at different layers disentangle different categories of meaning; and that different training techniques can have a significant effect on the interpretability of the representation learned by hidden units.

\ifCLASSOPTIONcompsoc
  \section*{Acknowledgments}
\else
  \section*{Acknowledgment}
\fi

This work was funded by DARPA XAI program No. FA8750-18-C-0004 and NSF Grant No. 1524817 to A.T., NSF grant No. 1532591 to A.O and A.T.; the Vannevar Bush Faculty Fellowship program funded by the ONR grant No. N00014-16-1-3116 to A.O.; the MIT Big Data Initiative at CSAIL, the Toyota Research Institute MIT CSAIL Joint Research Center, Google and Amazon Awards, and a hardware donation from NVIDIA Corporation. B.Z. is supported by a Facebook Fellowship.

\ifCLASSOPTIONcaptionsoff
  \newpage
\fi



% \bibliographystyle{IEEEtran}

{
\bibliographystyle{IEEEtran}
\bibliography{mainbib}
}
% \normalsize


\begin{IEEEbiography}[{\includegraphics[width=1in,height=1.25in,clip,keepaspectratio]{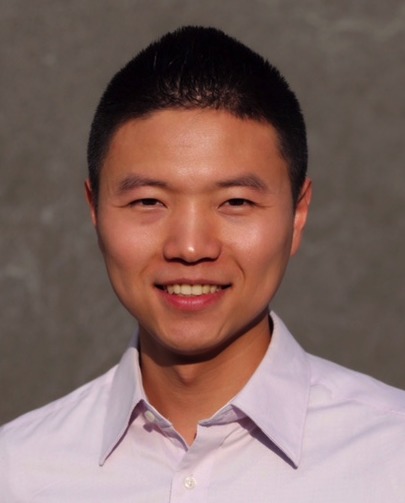}}]{Bolei Zhou}
is a Ph.D. Candidate in Computer Science and Artificial Intelligence Lab (CSAIL) at the Massachusetts Institute of Technology. He received M.Phil. degree in Information Engineering from the Chinese University of Hong Kong and B.Eng. degree  in Biomedical Engineering from Shanghai Jiao Tong University in 2010. His research interests are computer vision and machine learning. He is an award recipient of Facebook Fellowship, Microsoft Research Asia Fellowship, and MIT Greater China Fellowship.
\end{IEEEbiography}
\vskip -20pt plus -1fil
\begin{IEEEbiography}[{\includegraphics[width=1in,height=1.25in,clip,keepaspectratio]{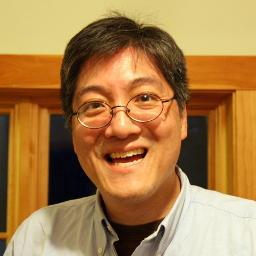}}]{David Bau}
is a PhD student at the MIT Computer Science and Artificial Intelligence Laboratory (CSAIL). He received an A.B. in Mathematics from Harvard in 1992 and an M.S. in Computer Science from Cornell in 1994. He coauthored a textbook on numerical linear algebra. He was a software engineer at Microsoft and Google and developed ranking algorithms for Google Image Search.  His research interest is interpretable machine learning.
\end{IEEEbiography}
\vskip -20pt plus -1fil
\begin{IEEEbiography}[{\includegraphics[width=1in,height=1.25in,clip,keepaspectratio]{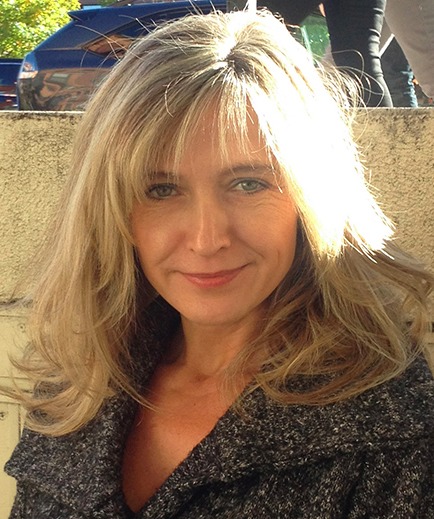}}]{Aude Oliva}
is a Principal Research Scientist at the MIT Computer Science and Artificial Intelligence Laboratory. After a baccalaureate in Physics and Mathematics, she received M.Sc and Ph.D degrees in Cognitive Sciences from the Institut National Polytechnique of Grenoble, France. She received the 2006 National Science Foundation Career award, the 2014 Guggenheim
and the 2016 Vannevar Bush awards. Her research spans cognitive science, neuroscience and computer vision.
\end{IEEEbiography}
\vskip -20pt plus -1fil
\begin{IEEEbiography}[{\includegraphics[width=1in,height=1.25in,clip,keepaspectratio]{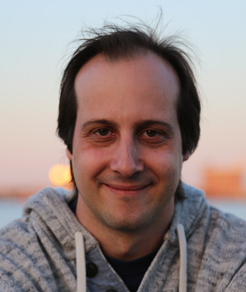}}]{Antonio Torralba}
received the degree in telecommunications engineering from Telecom BCN, Spain, in 1994 and the Ph.D. degree in signal, image, and speech processing from the Institut National Polytechnique de Grenoble, France, in 2000. From 2000 to 2005, he spent postdoctoral training at the Brain and Cognitive Science Department and the Computer Science and Artificial Intelligence Laboratory, MIT.  He is now a Professor of Electrical Engineering and Computer Science at the Massachusetts Institute of Technology (MIT).  Prof. Torralba is an Associate Editor of the International Journal in Computer Vision, and has served as program chair for the Computer Vision and Pattern Recognition conference in 2015. He received the 2008 National Science Foundation (NSF) Career award, the best student paper award at the IEEE Conference on Computer Vision and Pattern Recognition (CVPR) in 2009, and the 2010 J. K. Aggarwal Prize from the International Association for Pattern Recognition (IAPR).
\end{IEEEbiography}


%\vfill

% \enlargethispage{-1in}

%\appendices
%\section{}

% \begin{figure}
% \vspace{-5mm}
% \begin{center}
% \includegraphics[width=1\linewidth]{plot/journal_deepfeature-eps-converted-to.pdf}
% \end{center}
% \vspace{-5mm}
% \caption{The classification accuracy of deep features on the six image datasets.}\label{classification_accuracy}
% \end{figure}

\end{document}